\def\optionworldarxiv{1}
\documentclass[11pt]{article}
\ifdefined\pdfminorversion\pdfminorversion=7\fi

\ifdefined\optionworldarxiv
\usepackage[preprint]{acl}
\else
\usepackage[review]{acl}
\fi
\usepackage{stfloats}
\usepackage{times,latexsym}
\usepackage[T1]{fontenc}
\usepackage[utf8]{inputenc}
\usepackage{inconsolata}
\usepackage{amsmath,amssymb,mathtools}
\usepackage{booktabs,tabularx,multirow,microtype}
\usepackage{colortbl}
\usepackage{enumitem}
\usepackage{graphicx}
\newcommand{\ow}{\textsc{OptionWorld}}
\newcommand{\opr}{\mathrm{OPR}}

\title{OptionWorld: A Benchmark for Diagnosing Failures of LLMs\\
as Long-Term Personal Assistants}
\ifdefined\optionworldarxiv
\author{%
\begin{tabular}{@{}c@{\hspace{1.2em}}c@{\hspace{1.2em}}c@{}}
{\normalsize\bfseries Ruoyu Zhang$^{1}$} &
{\normalsize\bfseries Zhaocen Liu$^{1}$} &
{\normalsize\bfseries Weiyi Tian$^{2}$}\\[-0.15em]
{\footnotesize\normalfont\ttfamily zhangruoy@outlook.com} &
{\footnotesize\normalfont\ttfamily liu0411@student.ubc.ca} &
{\footnotesize\normalfont\ttfamily weiyitian@uchicago.edu} \\
\end{tabular}\\[0.35em]
\begin{tabular}{@{}c@{\hspace{1.8em}}c@{}}
{\normalsize\bfseries Melissa Feng$^{3}$} &
{\normalsize\bfseries Shilong Liu$^{4,\ast}$}\\[-0.15em]
{\footnotesize\normalfont\ttfamily mxf4246@nyu.edu} &
{\footnotesize\normalfont\ttfamily shilong.liu@princeton.edu}\\
\end{tabular}\\[0.65em]
{\small\normalfont $^{1}$University of British Columbia \quad $^{2}$University of Chicago}\\
{\small\normalfont $^{3}$Independent Researcher \quad $^{4}$Princeton University \quad $^{\ast}$Corresponding author}}

\hypersetup{pdfauthor={Ruoyu Zhang, Zhaocen Liu, Weiyi Tian, Melissa Feng, Shilong Liu},
  pdftitle={OptionWorld: A Benchmark for Diagnosing Failures of LLMs as Long-Term Personal Assistants}}
\else
\author{Anonymous authors}
\fi

\begin{document}
\maketitle
\raggedbottom
\begin{abstract}
Long-term assistants must follow unfinished tasks across days as information
and circumstances change. Finding information or starting an action,
however, does not ensure that the task gets completed. We introduce
\ow{}, a benchmark that traces information acquisition, judgments,
actions, and outcomes in simulated daily-life tasks to identify where
current LLMs fall short as long-term personal assistants. In runs with
archived decision inputs, the key facts remain in the model's last input
before the deadline in 35.5\% of failed instances. Failure analysis
identifies the three most frequent problems: needed information is not
obtained or combined, available opportunities to act go unused, and agents
impose restrictions that the task does not require. Three follow-up
experiments show that supplying missing facts restores task success in
some cases; action reminders can restart work while subsequent steps still
miss the deadline; and correcting a supplied note that wrongly forbids
contact restores permitted inquiry. These findings locate shortcomings
in maintaining task state: connecting facts to unfinished work, following
through before deadlines, and separating actual restrictions from
the agent's assumptions.
\end{abstract}

\section{Introduction}

\begin{figure*}[t]
\centering
\includegraphics[width=\textwidth,trim=0 20bp 0 0,clip]{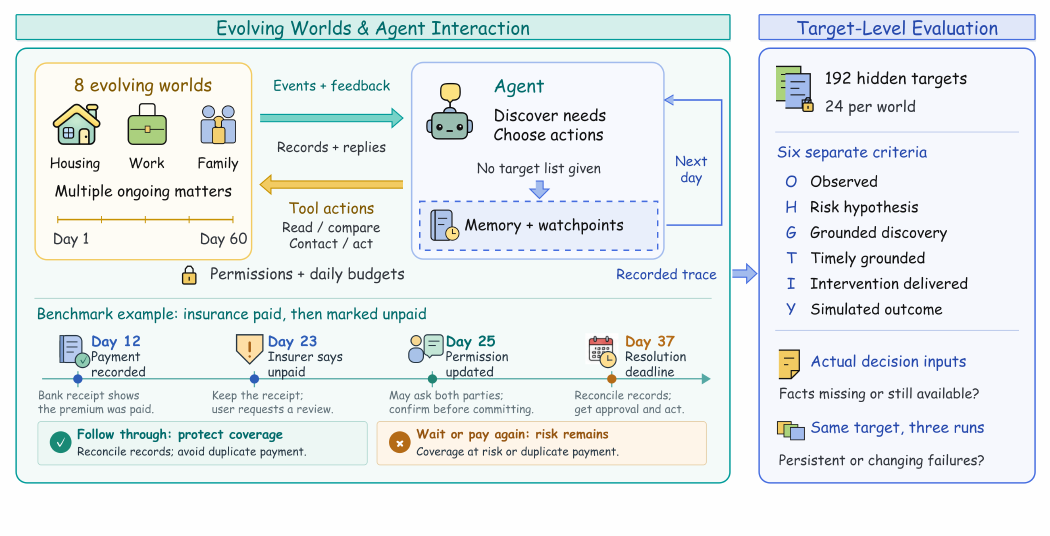}
\caption{\textbf{Following each target from evidence to action and outcome.} Agents discover needs through
events and tools, carrying memory and watchpoints across days.
In the insurance example, a paid premium is later listed as outstanding.
Updated permission allows consultation, but final commitments still require
confirmation; the agent must complete the protective action by Day~37.
Hidden-target evaluation links six judgments per target to decision inputs
and repeated outcomes. Six systems in eight 60-day worlds yield
144 trajectories over three runs.}
\label{fig:framework}
\vspace{-4pt}
\end{figure*}

\textbf{The problem: sustaining follow-through.}
LLM agents use language models to choose actions through tools, such as
reading records or contacting people \citep{yao2023react}.
Agents also retain memories and reflections
\citep{packer2023memgpt,shinn2023reflexion}.
As long-term personal assistants,
they must follow unfinished work as information, deadlines, and user needs
change. For example, an insurer may report an unpaid premium despite an
earlier bank receipt (Figure~\ref{fig:framework}). The assistant must connect
the records, follow updated permission to consult both parties, and complete
the user-approved action before the deadline. Maintaining
\emph{task state} means tracking unfinished work, permitted next steps, and
their deadlines. We ask where current LLMs fall short in this responsibility
and what support helps them resume progress.

\textbf{Our approach: locating where tasks stall.}
Existing benchmarks examine task execution, memory, and proactive
assistance \citep{zhou2024webarena,wu2024longmemeval,
wu2026longmemevalv2,pasternak2025probe,vibelifebench2026}.
Task and stage scores alone, however, leave open whether a failed target
lacked necessary facts, retained those facts without using them effectively,
or stalled during subsequent action. Distinguishing these problems requires
connecting outcomes to the information and decisions behind them. We introduce
\ow{}, a benchmark of simulated insurance, work, housing, and other
ongoing matters. Agents discover needs through daily events and records
without seeing the evaluation targets. For each target, we connect fact
acquisition, the facts still visible at decision time, judgments, actions,
and outcomes across days to locate where progress stops.

\textbf{Main findings: information is only part of the problem.}
The three most frequent problems concern connecting facts to unfinished
work and updating next steps: missing or unassembled
information, unused follow-ups or retries, and restrictions the agent adds
without a basis in the task. The fact-acquisition analysis finds that
nearly half of failures occur after
the primary facts were obtained in time. In runs with archived inputs,
over one-third of failures still have those facts in the last valid
decision input. Failure therefore extends beyond information acquisition.
Cross-day records help explain how obstacles develop: one agent saves a
rejection as a requirement to wait for another record and continues waiting
days later, although the rejection never imposed that condition. Here,
memory preserves a mistaken interpretation that guides later inaction.

Exploratory follow-ups test whether targeted support restores progress:
adding missing facts recovers some tasks, while reminders can restart an
action without completing the remaining work in time.
A controlled permission-note experiment further shows that correcting a
supplied false contact ban restores permitted inquiry while actual
no-contact rules remain respected.

\textbf{The main contributions of this work:}
\begin{itemize}[nosep]
\item A \textbf{benchmark} testing whether
assistants discover needs and complete tasks across days under changing
information and permissions.
\item A \textbf{target-level analysis} exposing unused
follow-ups and unsupported restrictions preserved in memory despite
available key facts.
\item Three \textbf{support experiments} distinguishing recovery of a
permitted action from completion of the original task.
\end{itemize}

\section{Related Work}

\providecommand{\owfull}{\textcolor{green!50!black}{$\checkmark$}}
\providecommand{\owpartial}{\textcolor{orange!85!black}{$\triangle$}}
\providecommand{\ownone}{\textcolor{red!70!black}{$\times$}}
\begin{table*}[t]
\centering
\footnotesize
\setlength{\tabcolsep}{2.8pt}
\renewcommand{\arraystretch}{1.18}

\begin{tabular}{@{}p{3.2cm}cccccccc@{}}
\toprule
& \multicolumn{4}{c}{\textbf{Agent capabilities}}
& \multicolumn{4}{c}{\textbf{Failure analysis}} \\
\cmidrule(lr){2-5}
\cmidrule(lr){6-9}

\textbf{Benchmark}
& \shortstack{\textbf{Need}\\\textbf{discovery}}
& \shortstack{\textbf{State}\\\textbf{tracking}}
& \shortstack{\textbf{Active}\\\textbf{search}}
& \shortstack{\textbf{Constraints}\\\textbf{\& deadlines}}
& \shortstack{\textbf{Access vs.}\\\textbf{judgment}}
& \shortstack{\textbf{Judgment--}\\\textbf{outcome link}}
& \shortstack{\textbf{Decision-input}\\\textbf{facts}}
& \shortstack{\textbf{Repeated}\\\textbf{success}} \\
\midrule

$\tau$-bench \citeyearpar{yao2024taubench}
& \ownone & \owpartial & \owfull & \owpartial
& \ownone & \owpartial & \ownone & \owfull \\

LongMemEvalV2 \citeyearpar{wu2026longmemevalv2}
& \ownone & \owfull & \owpartial & \ownone
& \owpartial & \ownone & \owpartial & \ownone \\

PM-Bench \citeyearpar{liu2026pmbench}
& \ownone & \owfull & \owfull & \owpartial
& \owpartial & \owpartial & \owpartial & \ownone \\

ChronosBench \citeyearpar{shi2026chronosbench}
& \ownone & \owfull & \owfull & \owpartial
& \ownone & \owpartial & \ownone & \ownone \\

VitaBench 2.0 \citeyearpar{chen2026vitabench}
& \owpartial & \owfull & \owfull & \owpartial
& \owpartial & \owpartial & \ownone & \owfull \\

$\pi$-Bench \citeyearpar{zhang2026pibench}
& \owfull & \owpartial & \owfull & \owpartial
& \owpartial & \owpartial & \ownone & \owpartial \\

PROBE \citeyearpar{pasternak2025probe}
& \owfull & \ownone & \owfull & \owpartial
& \owfull & \owpartial & \owpartial & \ownone \\

Claw-Anything \citeyearpar{lin2026clawanything}
& \owpartial & \owfull & \owfull & \ownone
& \owpartial & \owpartial & \owpartial & \owfull \\

VibeLifeBench \citeyearpar{vibelifebench2026}
& \owpartial & \owfull & \owfull & \owfull
& \owpartial & \owpartial & \ownone & \owpartial \\

\midrule
\rowcolor{gray!8}
\textbf{\ow{} (ours)}
& \owfull & \owfull & \owfull & \owfull
& \owfull & \owfull & \owfull & \owfull \\
\bottomrule
\end{tabular}

\caption{\textbf{Comparison of benchmark capabilities and failure analysis.}
\owfull{} denotes explicit coverage, \owpartial{} partial or related
coverage, and \ownone{} no reported coverage.
Discovery includes unstated needs and hidden requirements.
Constraints require both deadlines and permissions (one alone: \owpartial{}).
Failure analysis is target-level. Criteria and evidence:
Appendix~\ref{sec:comparison-detail}.}
\label{tab:comparison}
\vspace{-5pt}
\end{table*}

\paragraph{Agent methods and benchmarks.}
Generative Agents uses memory-based planning \citep{park2023generative},
MemGPT manages memory tiers \citep{packer2023memgpt}, and Voyager accumulates
reusable skills \citep{wang2023voyager}.
To evaluate agents, task-execution benchmarks ask whether they can complete a request
\citep{liu2023agentbench,zhou2024webarena,xie2024osworld,yao2024taubench}.
Memory benchmarks ask whether agents can recall and update information from
earlier interactions
\citep{wu2024longmemeval,wu2026longmemevalv2}.
Proactive-assistance benchmarks ask when to offer help and how to uncover
users' unstated needs or preferences
\citep{liu2025proactiveeval,tang2026proagentbench,zhang2026pibench,chen2026vitabench},
while other benchmarks test whether agents notice later changes or remember
to act when needed \citep{shi2026chronosbench,liu2026pmbench}.
These abilities support discovering and following up on users' needs
(Table~\ref{tab:comparison}, left). We study how agents combine them across
days: why do they stop following up or miss deadlines as information changes?

\paragraph{Locating failures and testing support.}
AgentBoard measures intermediate progress to analyze agent behavior beyond
final success \citep{ma2024agentboard}.
PROBE separately scores finding documents, identifying the problem, and
specifying an action \citep{pasternak2025probe}. VibeLifeBench checks behavior
and task state across multiple weeks \citep{vibelifebench2026}.
Claw-Anything tests how access to events and the amount of context affect
performance \citep{lin2026clawanything}.
Repeated-success measures check whether agents complete the same tasks
consistently across runs \citep{yao2024taubench,lin2026clawanything}.
\ow{} links these questions: which targets repeatedly fail, and where does
progress stop within them?
We connect source access and problem recognition to each target's key facts
at decision time, subsequent actions, and outcome feedback
(Table~\ref{tab:comparison}, right).

\section{Benchmark and Evaluation Design}

\subsection{Worlds and Interaction}

\ow{} contains eight authored daily-life simulations spanning housing, work,
independent living, and family coordination. A \emph{target} is a specific
issue to discover and address. Each 60-day world contains 24 targets,
totaling 192. Agents receive broad responsibilities, so they must discover
these needs from events and records without seeing the target list.
Evaluator-only specifications define required evidence, response periods,
and success conditions. Tasks may require comparing records across sources
or days, noticing that expected events have not occurred, or seeking
clarification or authorization.

\looseness=1
Daily inputs contain events, lists of available records, saved notes,
registered monitoring conditions, and feedback. Tools let the agent inspect
and compare records, save notes, contact relevant parties, request approval,
and carry out authorized operations. Visible rules specify what is allowed
and whose approval is required. Delayed replies may resolve only part of an
issue, requiring further updates and \mbox{follow-up (Figure~\ref{fig:framework})}.

\begingroup
\renewcommand{\arraystretch}{1.2}
\begin{table*}[t]
\centering\small\setlength{\tabcolsep}{3pt}
\begin{tabular*}{\textwidth}{@{\extracolsep{\fill}}lrrrrrr@{}}
\toprule
System & \shortstack[r]{Source\\access ($O$)} & \shortstack[r]{Risk\\hypothesis ($H$)} & \shortstack[r]{Grounded\\discovery ($G$)} & \shortstack[r]{Timely\\grounded ($T$)} & \shortstack[r]{Intervention\\delivered ($I$)} & \shortstack[r]{Target success\\($Y$ / OPR)} \\
\midrule
DeepSeek & 96.5\% & 86.3\% & 80.6\% & 78.1\% & 57.1\% & {\boldmath$48.3\% \pm 13.0\%$} \\
Fable & 84.9\% & 60.2\% & 55.0\% & 54.9\% & 40.6\% & {\boldmath$36.1\% \pm 5.2\%$} \\
Gemini & 87.5\% & 56.2\% & 49.7\% & 48.1\% & 37.5\% & {\boldmath$30.7\% \pm 6.6\%$} \\
Grok & 86.8\% & 58.9\% & 50.7\% & 50.2\% & 28.0\% & {\boldmath$23.6\% \pm 4.5\%$} \\
Qwen & 81.4\% & 63.0\% & 54.7\% & 54.3\% & 40.8\% & {\boldmath$35.8\% \pm 4.9\%$} \\
Astra (Codex) & 85.1\% & 80.7\% & 78.5\% & 77.8\% & 68.6\% & {\boldmath$61.6\% \pm 2.6\%$} \\
\bottomrule
\end{tabular*}
\caption{Main results: mean percentages across three runs (192 targets each); $\pm$ denotes sample standard deviation across runs. OPR: option-preservation rate (target success). Per-run counts: Table~\ref{tab:all-seed-stages}.}
\label{tab:mean-stages}
\end{table*}

\endgroup

\subsection{Evidence-to-Outcome Criteria}
\label{sec:criteria}

To locate where progress stops, we score the outcome and the preceding
evidence, judgments, and actions. Success requires feedback confirming that
the agent preserved a choice the user would otherwise lose or resolved the
issue. For example, confirmed protection of a pending clinic pickup
preserves an option; an accepted inquiry alone does not establish that outcome.

Each target receives six binary scores per run:
\textbf{Observed ($O$).} The agent accessed a designated relevant source
or record; this does not necessarily provide all required facts.
For example, a source listing does not show the contents of each record.
\textbf{Risk hypothesis ($H$).} It identified the specific problem or
risk associated with the target.
\textbf{Grounded discovery ($G$).} Its judgment was supported by the
required evidence already obtained in the interaction.
\textbf{Timely grounded discovery ($T$).} Its supported judgment fell
within the target's response period.
\textbf{Intervention delivered ($I$).} A target-specific external
action received an acceptance receipt and met the requirements for
evidence, timing, and permission specified for that target.
\textbf{Simulated outcome ($Y$).} Feedback linked to that action and
target confirmed that the user's option was preserved or the issue resolved.

Scores report confirmed credit. Unresolved review judgments retain a
separate uncertainty flag, reported as \emph{unknown} in the follow-ups.
Our main score, the option-preservation rate (OPR), is the success fraction:
$\opr=K^{-1}\sum_{r=1}^{K}Y_r$. Each target in each run counts as one instance;
$K$ is their total number.

Separating source access ($O$) from problem recognition ($H$), and accepted
intervention ($I$) from outcome ($Y$), locates different breaks in progress.
Scores satisfy $H\geq G\geq T\geq I\geq Y$: later criteria
require the earlier ones. These dependencies govern scoring, not action order.

\subsection{Evaluation Protocol}
\label{sec:cohorts}

We evaluate DeepSeek, Fable, Gemini, Grok, Qwen, and Astra (Codex) on
the same eight worlds, with three runs per world: 144 trajectories and
3,456 target instances in total. Five systems use the same runner, which
starts a new conversation each day while carrying forward saved notes,
monitoring conditions, and feedback; previously accessed records remain
retrievable. Astra uses a separate local Codex execution setup.
In runs with archived inputs, we check whether each failed target's primary
facts remain in its last decision input within the allowed action period,
distinguishing earlier acquisition from availability at decision time.

All trajectories use the same scoring rules. Two LLM reviewers independently
inspect records with the evaluated system's identity hidden. Their judgments
are combined with checks of action timing, acceptance receipts, and outcome
feedback. Multiple human reviewers also examined all 144 trajectories.
Appendices~\ref{sec:construction}--\ref{sec:review-process} give world
construction, execution settings, prompts, and scoring examples.

\section{Main Results and Failure Analysis}
\label{sec:results}

\subsection{Task Completion \& Where Progress Stops}

Under the specified environment budgets, Table~\ref{tab:mean-stages} reports
target success from 23.6\% to 61.6\%:
even the highest-scoring system leaves more than one-third unfinished.
Astra has the highest success rate (61.6\%), followed by DeepSeek (48.3\%).

\paragraph{Broader source access does not ensure more completed tasks.}
\looseness=-1
Astra has lower source-access coverage than DeepSeek (85.1\% versus 96.5\%)
and has similar timely grounded discovery (77.8\% versus 78.1\%), yet
delivers more interventions (68.6\% versus 57.1\%) and completes more
targets. Source access records whether at least one relevant source was
accessed, rather than the amount of information gathered. To compare
follow-through on the same tasks, we match 275 instances for which both
systems make timely grounded judgments and have no uncertain stage labels.
Astra delivers 268 interventions and completes 262 targets; DeepSeek
delivers 227 and completes 207.

In a paired trajectory, both systems identify conflicting weekly
availability requirements.
Astra obtains authorization and submits a protective correction request,
leaving the factual reconciliation to the responsible record holder;
DeepSeek saves the discrepancy and waits for that holder to become
available. Recognized uncertainty can thus lead either to a permitted
protective step or to continued monitoring
(Appendix~\ref{sec:astra-follow-through}).

\paragraph{Recognizing a problem can still leave the next step undone.}
Grok makes timely grounded judgments for 50.2\% of targets but delivers
interventions for only 28.0\%, ending with the lowest success rate (23.6\%).
Of its 289 timely judgments, 128 do not lead to a qualifying intervention.
The bottleneck differs from Gemini's earlier gap: Gemini accesses sources
for 87.5\% of targets but identifies a specific risk for only 56.2\%.
Even among 239 matched instances with timely primary facts and no uncertain
stage judgments for either DeepSeek or Grok, they deliver 185 and 127
interventions, respectively.
In a supplier-payment case, both obtain a credit memo and a conflicting
full-debit record. Grok records the conflict and a watchpoint; DeepSeek
obtains confirmation and executes the approved action before the deadline
(Appendices~\ref{sec:paths} and~\ref{sec:paired-actions}).

\begin{figure*}[t]
\centering
\includegraphics[width=\textwidth]{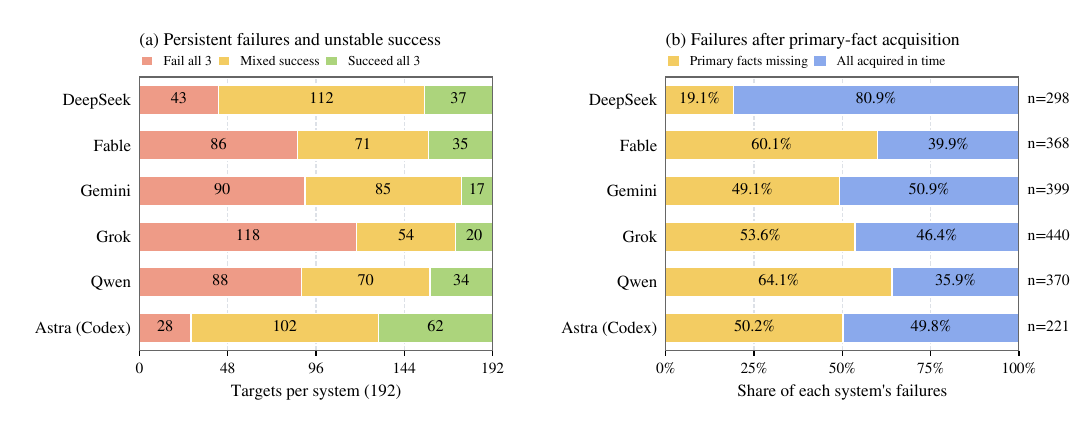}
\caption{\textbf{Repeated outcomes and fact acquisition.}
Six systems, three runs. (a) Outcomes across runs for 192 targets per system;
mixed success means succeeding once or twice. (b) Failed run--target instances
grouped by whether all designated primary facts were acquired before the
deadline; $n$ denotes each system's failure count.}
\label{fig:main-findings}
\vspace{-4pt}
\end{figure*}

\label{sec:acquisition}
\label{sec:context}
\looseness=-1
Could these gaps instead reflect missing or forgotten facts?
Across all six systems and three runs,
1,038 of 2,096 failures (49.5\%) occur after the agent
obtains all designated primary facts before the deadline
(Figure~\ref{fig:main-findings}b). For the five systems sharing a runner,
the input-visibility analysis uses only Runs 2--3, where decision inputs are
archived: 460 of their 1,295 failures (35.5\%)
still contain those facts in the last input within the allowed action
period (Appendix~\ref{sec:context-audit}, Table~\ref{tab:context}).

\paragraph{Similar scores conceal different successes and activity.}
Fable and Qwen complete 208 and 206 of 576 instances, respectively, but
these are not the same successful tasks. They disagree on 150 matched
outcomes: 76 favor Fable and 74 favor Qwen. The difference remains on 78
of 249 matched instances where both obtain the primary facts in time and
neither has uncertain stage labels.
The contrast runs in both directions within the same caregiver world and
run, where both systems complete 12 of 24 targets. Fable follows up an unanswered
authorization request and completes a care-allowance task; Qwen's attempts
stall after rejected requests. On a childcare funding discrepancy, Qwen
follows a partial reply through to clarification and action, while Fable
waits for another record before asking. The shared facts still allow
different choices: inquire, repair a rejected request, or wait
(Appendix~\ref{sec:paired-actions}).

Nor does Qwen's greater activity produce a higher total: it makes 12,553
tool calls versus Fable's 8,049 across the same 24 world--run pairs.
Of the 4,504 additional calls, 1,967 (43.7\%) are risk reports or watchpoint
registrations. Extra activity can thus record and monitor pending matters
without acting to resolve them
(Appendix~\ref{sec:resource-observation}).

\paragraph{Fewer records need not mean easier completion.}
Among shared-runner instances with timely grounded judgments and no uncertain
stage labels, targets involving an expected event that has not occurred
(missing-event) succeed in 90.7\% of cases, compared with 68.4\% for
constraints stated in one record (single-source) and 60.2\% for tasks
requiring facts from different days (cross-day accumulation).
All instances in this comparison have
timely primary facts, so the difference remains after discovery.
Missing-event tasks proceed through a request to the responsible
institution, whereas cross-day tasks require confirmation followed by
execution. A task understood from one record is therefore not necessarily
easier to complete than one requiring several records: the subsequent
action requirements also matter (Appendix~\ref{sec:task-paths}).

\paragraph{Repeated runs reveal persistent failures and unstable success.}
\label{sec:reliability}
Do repeated runs overcome the same obstacles? Across all six
systems, 64.8\% of failed instances come from targets that the same system
fails in all three runs (Figure~\ref{fig:main-findings}a). In the shared-runner
group, of 425 model--target pairs failing all three runs, 305 concern
targets another system completes at least once.
Only 24 of 192 targets fail across all 15 attempts. Persistent failure
within one system therefore need not be shared by the others.

\looseness=-1
Among 301 model--target pairs in this group with timely primary facts, timely grounded judgment,
and no uncertain stage labels in any of the three runs, 119 succeed only
once or twice and 39 never succeed. Stable discovery does not ensure
stable completion. For example, DeepSeek identifies the
same pension-record conflict in all three runs. In two it takes a reversible
step to preserve the recorded service history; in the third it waits for
another record and never delivers an intervention. The repeated difficulty
lies in choosing and carrying out an available next step after discovery
(Appendices~\ref{sec:sensitivity} and~\ref{sec:paired-actions}).
We next trace these failures.

\subsection{Failure Profiles and Trajectory Evidence}
\label{sec:failure-profiles}

\begin{table*}[t]
\centering\small
\setlength{\tabcolsep}{4pt}
\centering\small
\begin{tabular*}{\linewidth}{@{\extracolsep{\fill}}lr>{\columncolor[HTML]{9FC5F3}}r>{\columncolor[HTML]{FFD56B}}r>{\columncolor[HTML]{A8D58E}}rrrr@{}}
\toprule
System & $N$ & \shortstack{Information\\gap} & \shortstack{Action\\selection /\\adjustment} & \shortstack{Self-\\restriction} & \shortstack{State /\\update\\error} & \shortstack{Execution\\error} & \shortstack{Insufficient\\evidence} \\
\midrule
DeepSeek & 296 & 18.6\% & 16.2\% & 20.6\% & 13.9\% & 1.7\% & 29.1\% \\
Fable & 367 & 46.3\% & 10.1\% & 23.4\% & 4.9\% & 0.3\% & 15.0\% \\
Gemini & 364 & 35.2\% & 23.4\% & 9.3\% & 6.3\% & 2.7\% & 23.1\% \\
Grok & 373 & 13.7\% & 47.7\% & 22.3\% & 3.8\% & 0.8\% & 11.8\% \\
Qwen & 325 & 24.9\% & 12.9\% & 29.8\% & 7.4\% & 6.2\% & 18.8\% \\
Astra (Codex) & 211 & 46.9\% & 10.4\% & 2.4\% & 1.4\% & 0.0\% & 38.9\% \\
\midrule
Total & 1936 & 30.2\% & 21.3\% & 18.9\% & 6.4\% & 2.0\% & 21.3\% \\
\bottomrule
\end{tabular*}
\caption{Primary-problem profiles by system ($N=1{,}936$); one label per instance. Rows total 100\% before rounding, including the insufficient-evidence status. Information gap: missing, unassembled, or unretained facts.}
\label{tab:failure-census}
\vspace{-4pt}
\end{table*}

The three most frequent problems are missing or unassembled
information (30.2\%), failure to take or adjust an available action
(21.3\%), and self-imposed restrictions (18.9\%).
Across all six systems, 2,096 instances are unsuccessful. We exclude 160
involving normal constraints alone, such as quota or permission limits, and
classify the remaining 1,936 instances (Table~\ref{tab:failure-census}).
Each receives one label; insufficient evidence (21.3\%) is a separate status
for instances where the record does not establish a specific problem.
Human reviewers checked every LLM-proposed classification against the
original trajectory and resolved disagreements before finalizing the labels
(Appendix~\ref{sec:failure-census}).
\par

Among Fable's and Qwen's classified instances,
information gaps account for 46.3\% and 24.9\%, while self-restriction
accounts for 23.4\% and 29.8\%. Their similar overall scores thus accompany
different failure profiles: Fable more often lacks needed information,
while Qwen more often imposes unsupported restrictions.
Grok's most frequent problem is action selection or adjustment (47.7\%
of its classified failures), consistent with its large gap between timely
judgment and intervention. This category includes choosing an ineffective
action or failing to adapt one, not only inaction.

\paragraph{Needed information is not obtained or assembled in time.}
The largest category includes facts never retrieved and facts read but not
combined or retained for later decisions. Missing facts can remain
accessible through record comparison after direct-inspection allowances
are exhausted. In a stratified review of 80 shared-runner failures,
35 lack primary facts at the deadline. Replaying comparison tools
at saved states confirms this alternative for 15 cases across four
systems. Five of these trajectories had already retrieved new facts by
comparison elsewhere. Those five expose
a strategy-reuse gap: an available, previously used retrieval method is not
applied to the unresolved matter after inspection quotas run out
(Appendix~\ref{sec:strategy-reuse}).

\paragraph{Available actions do not always become effective interventions.}
The second category concerns missed follow-ups, retries, or alternative
approaches when a permitted opportunity exists. In a Gemini food-business
trajectory, the agent notices overlapping inspection and pickup times and
tries to ask for clarification on Day 10, when the contact is unavailable.
By Day 15, the relevant parties are available; the input still contains both
time windows and the user's preference to keep both options open. Asking
for clarification is allowed, with time left before the Day-22 deadline,
but the agent does not return to the consultation. Task memory must support
reconsidering what to do when a temporary obstacle
disappears, not merely preserving the unfinished task
(Appendix~\ref{sec:recovery-protocol}).

A contrasting success pattern appears when feedback specifies how to
revise a request. When a request is returned for correction or only partly
resolved, the feedback names the field to correct and the required source
format. We examine instances
where both the feedback and its specified follow-up day fall within the
deadline. In the shared-runner group, all 99 eligible returned-material instances
and all 126 partial-resolution instances receive accepted follow-up
requests before the deadline; 91 and 109, respectively, reach the target
outcome. For example,
Grok rereads a licensing record, supplies the missing case reference with
its source, and obtains a resolution. These interactions demonstrate local
repair under explicit guidance. Sustained follow-through also requires
recognizing when to resume a pending task without such a specific revision
instruction, as the missed consultation above illustrates
(Appendix~\ref{sec:feedback-repair}).

\paragraph{The agent adds conditions that the task does not require.}
\label{sec:capabilities}

Self-restriction turns an unsupported assumption into a reason to wait or
stop. In one Qwen trajectory, a rent-review request on Day 19 is rejected
with a message that the evidence does not match the specified procedure.
Qwen conjectures that another record is missing and saves the instruction
``retry after a companion entry.''
On Day 24, it explicitly reports not retrying because no such entry has
arrived. The rejection never required that record
(Appendix~\ref{sec:restriction-excerpts}).

The same trajectory correctly waits for a daily quota to reset on another
matter and retries the next day. It can retry, but does not consistently
distinguish environmental requirements from its own explanations:
memory can perpetuate mistakes as well as preserve facts. Permission
errors similarly remove an allowed next step when a ban on commitment
is treated as a ban on inquiry.

\paragraph{Completing a procedure is not the same as resolving the task.}
In the shared-runner group, preference, authorization, and multi-party tasks
account for 18.2\% of target instances but 60.6\% of cases with a
qualifying intervention and no successful outcome; no such cases occur
among single-source, cross-source, or cross-day targets in these runs
\mbox{(Appendix Table~\ref{tab:stage-types})}.

Among 122 multi-party instances with timely facts and judgment and no uncertain
stage labels, both consultations, authorized execution, and tool acceptance
occur by the deadline, yet 28 fail to meet the target outcome.
In a matched DeepSeek mortgage case, both runs consult the parties and
execute an approved action. A hardship letter pauses an instalment, but the
bank still schedules its full debit. The successful run asks the responsible
office to apply a hardship pause; the failed run proposes treating the
hardship letter as authoritative and pausing the debit itself. The critical
difference is whether the action asks the appropriate party to reconcile
the conflict or assumes that the \mbox{conflict is already settled}.

\paragraph{Task state must track both unfinished and completed work.}
Less frequent problems are state or updating errors (6.4\%) and execution
errors (2.0\%). A scan of shared-runner trajectories shows that state
maintenance also requires stopping completed operations: 69 target instances
in 33 trajectories incur 106
repeated-operation rejections after an accepted completion. Forty of these
targets are successful under the rubric, so success rates alone miss this
redundant activity. The pattern is concentrated in Gemini (59 instances):
one trajectory executes on Day 6, receives completion feedback on Day 7,
then reuses the same authorization on Days 7, 9, and 13.
Together with the missed-retry case, this reveals two
directions of state error: leaving a temporary block in place and failing
to retire a completed step. Execution errors instead concern parameters, object bindings,
or operation \mbox{order that block an action}.

\section{Targeted Support for Stalled Tasks}
\label{sec:followups}

At selected checkpoints, we test three forms of support corresponding to the problems identified
in Section~\ref{sec:failure-profiles}: supplying missing facts, prompting
the next permitted action, and clarifying permission boundaries.
The first two experiments resume natural failures; the third tests permitted
inquiry under externally supplied incorrect permission notes and their corrections.

\begin{figure*}[t]
\centering
\includegraphics[width=\textwidth]{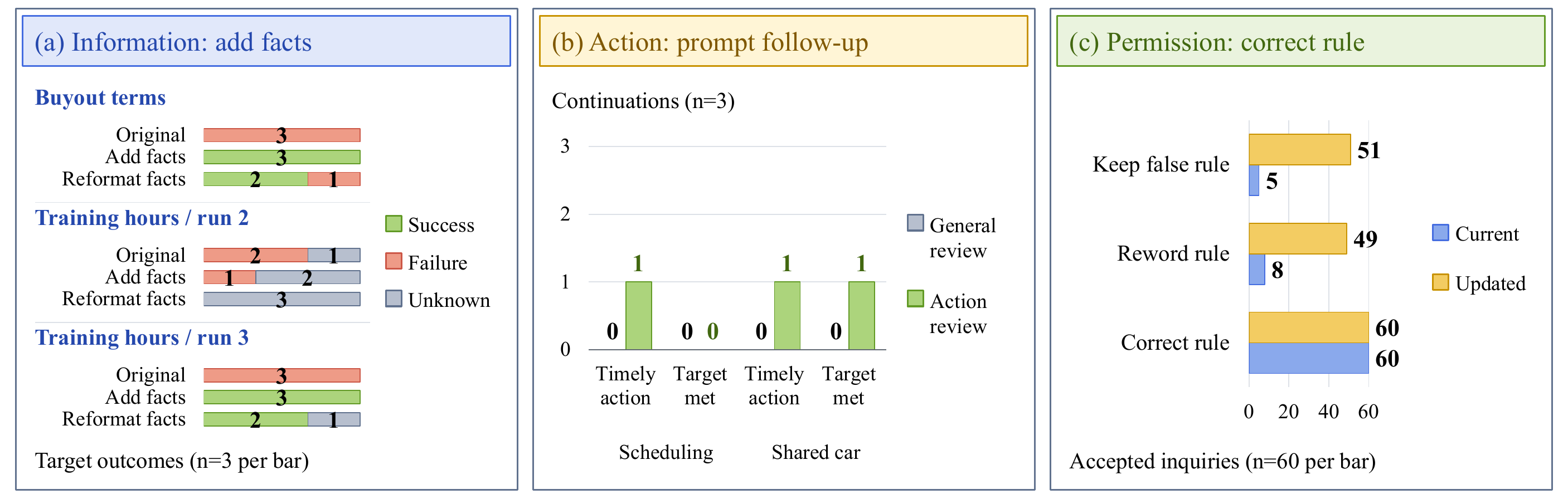}
\caption{\textbf{Targeted support experiments.}
(a--b) Three continuations per condition from natural failures.
(a) Target outcomes; reformatting transposes the fact table.
Both training checkpoints concern one target.
(b) Qualifying intervention ($I$) and target success ($Y$).
(c) Accepted inquiries under externally supplied permission notes:
six checkpoints $\times$ ten repeats. The updated presentation jointly
reorders records and releases waiting.}
\label{fig:followups}
\vspace{-4pt}
\end{figure*}

In the first two experiments, each condition resumes the saved world and
agent state (\emph{checkpoint}) in three \emph{continuations} through Day 60 with the
original model, permissions, budgets, and deadlines. We score new timely
intervention ($I$) and target outcome ($Y$) under the main-study rules,
retaining uncertain scores.
Figure~\ref{fig:followups} summarizes all three experiments;
Appendix~\ref{sec:recovery-protocol} gives the continuation protocol and full counts.

\subsection{Information: Supplying Missing Evidence}
\label{sec:information-experiment}

Three Gemini checkpoints cover a contractor's buyout and a worker's
training hours, the latter from original Runs 2 and 3. The buyout values
one campaign, but a ledger lists three renewals, requiring reconciliation
by the responsible record holder. The college grants 400 training hours,
while certification rules recognize 720 with supervisor proof.
In both cases, an allowed comparison can retrieve the missing facts.
We compare unchanged continuation (\emph{original}) with \emph{added facts}:
one researcher-selected comparison-tool call supplies its returned facts
and a table of both records. \emph{Reformatted facts} transposes the same
table without changing its contents. Neither supplies an interpretation
or action plan.

\paragraph{Added facts restore success at two checkpoints.}
\looseness=-1
At the buyout and Run-3 training checkpoints, added facts raise target
success from 0/3 to 3/3 (Figure~\ref{fig:followups}a). Reformatting yields
2/3 successes at each.
At Run-2 training, the two support conditions together yield one failure
and five uncertain outcomes across six continuations.
The uncertain outcomes concern whether the request seeks reconciliation
of 400 versus 720 hours or assumes eligibility without supervisory proof.

\subsection{Action: Reviewing the Next Step}
\label{sec:action-experiment}

For inspection--pickup scheduling (Gemini) and shared-car--clinic
coordination (DeepSeek), we compare two one-time reminders to review
unfinished work under existing permissions. Unlike \emph{general review},
\emph{action review} explicitly names follow-up,
retry after a lifted restriction, and another permitted approach.
Neither adds facts, resources, or a target-specific plan.

\paragraph{Restarting action and finishing in time are different requirements.}
Compared with general review, which yields 0/3 qualifying interventions in
each case, action review yields 1/3 (Figure~\ref{fig:followups}b). Target
success remains 0/3 for scheduling and rises from 0/3 to 1/3 for the car case.
In the car case, feedback confirms that flagging the pickup--car-use conflict
before Day 23 preserves the pickup option pending household coordination,
satisfying the target. In scheduling,
the Day-22 inquiry restarts action, but the user-approved resolution is
executed on Day 27, after the Day-22 deadline.

\subsection{Permission: Correcting a False Rule}
\label{sec:permission-experiment}

At one checkpoint in each of six worlds, Qwen3-Max receives an externally
supplied incorrect no-contact note: \emph{retain} the judgment,
\emph{rewrite} it with the same meaning, or \emph{correct} it to distinguish
inquiry from commitment. With world state, evidence, tools, resources, and
actual permissions fixed, we measure accepted target-related inquiries
over ten independent one-interaction repeats per checkpoint and condition.

\paragraph{Correcting the restriction restores permitted inquiry.}
When the note follows the observed records, correction yields 60/60 accepted
inquiries, versus 5/60 for retaining it and 8/60 for equivalent rewriting
(Figure~\ref{fig:followups}c).
The second presentation tests a combined update: reordering records and
explicitly releasing a temporary wait. Inquiry also recovers under this
update with the false note retained.
Across both presentations, 360 control
windows with genuine no-contact rules produce no target-contact attempts
(Appendix~\ref{sec:permission-protocol}).

\section{Discussion and Conclusion}

\ow{} locates shortcomings in long-term assistance by connecting each
unfinished target to its evidence, judgments, actions, and outcome.
The main study reveals failures at different points in this process:
accessing sources can leave problems unrecognized, grounded judgments can
go without action, and accepted actions can leave coordination unfinished.
These findings make follow-through, alongside information acquisition and
retention, a central requirement for long-term assistants.

Memory can preserve a mistaken reason to wait alongside useful facts,
connecting persistent early assumptions in dialogue \citep{laban2025lost}
to cross-day inaction. Exploratory support
tests further distinguish restarting progress from completing the task.
Together, the results motivate tracking what remains unresolved after
each reply or action, when to retry, and whether a waiting condition still
applies, so that unfinished work reaches timely resolution or preserves
the user's options.

\section*{Limitations}

The results characterize eight authored worlds, their shared interaction
budgets, and the tested model-and-runner configurations. Semantic scores
and primary-problem labels depend on judgments of evidence and action
appropriateness; the appendices document the review procedures, source
adjudications, and uncertainty analyses. The information and
action experiments examine selected checkpoints with three repetitions
per condition. Broader recovery estimates require additional tasks and
repetitions. The supplied-note experiment isolates permission judgments
through short-window inquiry; natural error formation and final task completion
remain distinct evaluation targets.

\section*{Ethical Considerations}

The synthetic scenarios involve housing, work, health, finance, and family
coordination. Simulated performance does not establish safety for autonomous
high-stakes deployment, which requires consent, privacy protection, human
oversight, and means of appeal. Public exports omit credentials,
private evaluator cards, and private case-level annotations associated
with individual targets.

\paragraph{AI assistance.}
AI assisted development,
orchestration, review, and drafting. Authors verify the sources and
take responsibility for all claims.

\clearpage
\bibliography{references}

@article{chen2026vitabench,
  author       = {Yuxin Chen and
                  Yi Zhang and
                  Zhengzhou Cai and
                  Yaorui Shi and
                  Zhiyuan Yao and
                  Chenhang Cui and
                  Jingnan Zheng and
                  Yaqi Huo and
                  Xi Su and
                  Qi Gu and
                  Xunliang Cai and
                  Xiang Wang and
                  An Zhang and
                  Tat{-}Seng Chua},
  title        = {VitaBench 2.0: Evaluating Personalized and Proactive Agents in Long-Term
                  User Interactions},
  journal      = {CoRR},
  volume       = {abs/2605.27141},
  year         = {2026},
  url          = {https://doi.org/10.48550/arXiv.2605.27141},
  doi          = {10.48550/ARXIV.2605.27141},
  eprinttype   = {arXiv},
  eprint       = {2605.27141},
  bibsource    = {dblp computer science bibliography, https://dblp.org}
}

@article{laban2025lost,
  author       = {Philippe Laban and
                  Hiroaki Hayashi and
                  Yingbo Zhou and
                  Jennifer Neville},
  title        = {LLMs Get Lost In Multi-Turn Conversation},
  journal      = {CoRR},
  volume       = {abs/2505.06120},
  year         = {2025},
  url          = {https://doi.org/10.48550/arXiv.2505.06120},
  doi          = {10.48550/ARXIV.2505.06120},
  eprinttype   = {arXiv},
  eprint       = {2505.06120},
  bibsource    = {dblp computer science bibliography, https://dblp.org}
}

@article{lin2026clawanything,
  author       = {Yusong Lin and
                  Xinyuan Liang and
                  Haiyang Wang and
                  Qipeng Gu and
                  Siqi Cheng and
                  Jiangui Chen and
                  Shuzhe Wu and
                  Feiyang Pan and
                  Lue Fan and
                  Sanyuan Zhao and
                  Dandan Tu},
  title        = {Claw-Anything: Benchmarking Always-On Personal Assistants with Broader
                  Access to User's Digital World},
  journal      = {CoRR},
  volume       = {abs/2605.26086},
  year         = {2026},
  url          = {https://doi.org/10.48550/arXiv.2605.26086},
  doi          = {10.48550/ARXIV.2605.26086},
  eprinttype   = {arXiv},
  eprint       = {2605.26086},
  bibsource    = {dblp computer science bibliography, https://dblp.org}
}

@article{liu2026pmbench,
  author       = {Genglin Liu and
                  Saadia Gabriel},
  title        = {PM-Bench: Evaluating Prospective Memory in {LLM} Agents},
  journal      = {CoRR},
  volume       = {abs/2607.12385},
  year         = {2026},
  url          = {https://doi.org/10.48550/arXiv.2607.12385},
  doi          = {10.48550/ARXIV.2607.12385},
  eprinttype   = {arXiv},
  eprint       = {2607.12385},
  bibsource    = {dblp computer science bibliography, https://dblp.org}
}

@inproceedings{liu2023agentbench,
  author       = {Xiao Liu and
                  Hao Yu and
                  Hanchen Zhang and
                  Yifan Xu and
                  Xuanyu Lei and
                  Hanyu Lai and
                  Yu Gu and
                  Hangliang Ding and
                  Kaiwen Men and
                  Kejuan Yang and
                  Shudan Zhang and
                  Xiang Deng and
                  Aohan Zeng and
                  Zhengxiao Du and
                  Chenhui Zhang and
                  Sheng Shen and
                  Tianjun Zhang and
                  Yu Su and
                  Huan Sun and
                  Minlie Huang and
                  Yuxiao Dong and
                  Jie Tang},
  title        = {AgentBench: Evaluating LLMs as Agents},
  booktitle    = {The Twelfth International Conference on Learning Representations,
                  {ICLR} 2024, Vienna, Austria, May 7-11, 2024},
  publisher    = {OpenReview.net},
  year         = {2024},
  url          = {https://openreview.net/forum?id=zAdUB0aCTQ},
  bibsource    = {dblp computer science bibliography, https://dblp.org}
}

@inproceedings{liu2025proactiveeval,
  author       = {Tianjian Liu and
                  Fanqi Wan and
                  Jiajian Guo and
                  Xiaojun Quan},
  editor       = {Maria Liakata and
                  Viviane P. Moreira and
                  Jiajun Zhang and
                  David Jurgens},
  title        = {ProactiveEval: {A} Unified Evaluation Framework for Proactive Dialogue
                  Agents},
  booktitle    = {Proceedings of the 64th Annual Meeting of the Association for Computational
                  Linguistics (Volume 1: Long Papers), {ACL} 2026, San Diego, California,
                  United States, July 2-7, 2026},
  pages        = {41068--41100},
  publisher    = {Association for Computational Linguistics},
  year         = {2026},
  url          = {https://doi.org/10.18653/v1/2026.acl-long.1906},
  doi          = {10.18653/V1/2026.ACL-LONG.1906},
  bibsource    = {dblp computer science bibliography, https://dblp.org}
}

@inproceedings{ma2024agentboard,
  author       = {Chang Ma and
                  Junlei Zhang and
                  Zhihao Zhu and
                  Cheng Yang and
                  Yujiu Yang and
                  Yaohui Jin and
                  Zhenzhong Lan and
                  Lingpeng Kong and
                  Junxian He},
  editor       = {Amir Globersons and
                  Lester Mackey and
                  Danielle Belgrave and
                  Angela Fan and
                  Ulrich Paquet and
                  Jakub M. Tomczak and
                  Cheng Zhang},
  title        = {AgentBoard: An Analytical Evaluation Board of Multi-turn {LLM} Agents},
  booktitle    = {Advances in Neural Information Processing Systems 37: Annual Conference
                  on Neural Information Processing Systems 2024, NeurIPS 2024, Vancouver,
                  BC, Canada, December 10 - 15, 2024},
  year         = {2024},
  url          = {http://papers.nips.cc/paper\_files/paper/2024/hash/877b40688e330a0e2a3fc24084208dfa-Abstract-Datasets\_and\_Benchmarks\_Track.html},
  bibsource    = {dblp computer science bibliography, https://dblp.org}
}

@article{packer2023memgpt,
  author       = {Charles Packer and
                  Vivian Fang and
                  Shishir G. Patil and
                  Kevin Lin and
                  Sarah Wooders and
                  Joseph E. Gonzalez},
  title        = {MemGPT: Towards LLMs as Operating Systems},
  journal      = {CoRR},
  volume       = {abs/2310.08560},
  year         = {2023},
  url          = {https://doi.org/10.48550/arXiv.2310.08560},
  doi          = {10.48550/ARXIV.2310.08560},
  eprinttype   = {arXiv},
  eprint       = {2310.08560},
  bibsource    = {dblp computer science bibliography, https://dblp.org}
}

@article{park2023generative,
  author       = {Joon Sung Park and
                  Joseph C. O'Brien and
                  Carrie J. Cai and
                  Meredith Ringel Morris and
                  Percy Liang and
                  Michael S. Bernstein},
  title        = {Generative Agents: Interactive Simulacra of Human Behavior},
  journal      = {CoRR},
  volume       = {abs/2304.03442},
  year         = {2023},
  url          = {https://doi.org/10.48550/arXiv.2304.03442},
  doi          = {10.48550/ARXIV.2304.03442},
  eprinttype   = {arXiv},
  eprint       = {2304.03442},
  bibsource    = {dblp computer science bibliography, https://dblp.org}
}

@article{pasternak2025probe,
  author       = {Gil Pasternak and
                  Dheeraj Rajagopal and
                  Julia White and
                  Dhruv Atreja and
                  Matthew Thomas and
                  George Hurn{-}Maloney and
                  Ash Lewis},
  title        = {Beyond Reactivity: Measuring Proactive Problem Solving in {LLM} Agents},
  journal      = {CoRR},
  volume       = {abs/2510.19771},
  year         = {2025},
  url          = {https://doi.org/10.48550/arXiv.2510.19771},
  doi          = {10.48550/ARXIV.2510.19771},
  eprinttype   = {arXiv},
  eprint       = {2510.19771},
  bibsource    = {dblp computer science bibliography, https://dblp.org}
}

@article{shi2026chronosbench,
  author       = {Qinglong Shi and
                  Donghai Wang and
                  Hantao Zhou and
                  Jiguo Li and
                  Jun Xu and
                  Jiuchong Gao and
                  Jinghua Hao and
                  Renqing He},
  title        = {Long-term Task-oriented Agent: Proactive Long-term Intent Maintenance
                  in Dynamic Environments},
  journal      = {CoRR},
  volume       = {abs/2601.09382},
  year         = {2026},
  url          = {https://doi.org/10.48550/arXiv.2601.09382},
  doi          = {10.48550/ARXIV.2601.09382},
  eprinttype   = {arXiv},
  eprint       = {2601.09382},
  bibsource    = {dblp computer science bibliography, https://dblp.org}
}

@inproceedings{shinn2023reflexion,
  author       = {Noah Shinn and
                  Federico Cassano and
                  Ashwin Gopinath and
                  Karthik Narasimhan and
                  Shunyu Yao},
  editor       = {Alice Oh and
                  Tristan Naumann and
                  Amir Globerson and
                  Kate Saenko and
                  Moritz Hardt and
                  Sergey Levine},
  title        = {Reflexion: language agents with verbal reinforcement learning},
  booktitle    = {Advances in Neural Information Processing Systems 36: Annual Conference
                  on Neural Information Processing Systems 2023, NeurIPS 2023, New Orleans,
                  LA, USA, December 10 - 16, 2023},
  year         = {2023},
  url          = {http://papers.nips.cc/paper\_files/paper/2023/hash/1b44b878bb782e6954cd888628510e90-Abstract-Conference.html},
  bibsource    = {dblp computer science bibliography, https://dblp.org}
}

@article{tang2026proagentbench,
  author       = {Yuanbo Tang and
                  Huaze Tang and
                  Tingyu Cao and
                  Lam Nguyen and
                  Anping Zhang and
                  Xinwen Cao and
                  Chunkang Liu and
                  Wenbo Ding and
                  Yang Li},
  title        = {ProAgentBench: Evaluating {LLM} Agents for Proactive Assistance with
                  Real-World Data},
  journal      = {CoRR},
  volume       = {abs/2602.04482},
  year         = {2026},
  url          = {https://doi.org/10.48550/arXiv.2602.04482},
  doi          = {10.48550/ARXIV.2602.04482},
  eprinttype   = {arXiv},
  eprint       = {2602.04482},
  bibsource    = {dblp computer science bibliography, https://dblp.org}
}

@article{wang2023voyager,
  author       = {Guanzhi Wang and
                  Yuqi Xie and
                  Yunfan Jiang and
                  Ajay Mandlekar and
                  Chaowei Xiao and
                  Yuke Zhu and
                  Linxi Fan and
                  Anima Anandkumar},
  title        = {Voyager: An Open-Ended Embodied Agent with Large Language Models},
  journal      = {Trans. Mach. Learn. Res.},
  volume       = {2024},
  year         = {2024},
  url          = {https://openreview.net/forum?id=ehfRiF0R3a},
  bibsource    = {dblp computer science bibliography, https://dblp.org}
}

@inproceedings{wu2024longmemeval,
  author       = {Di Wu and
                  Hongwei Wang and
                  Wenhao Yu and
                  Yuwei Zhang and
                  Kai{-}Wei Chang and
                  Dong Yu},
  title        = {LongMemEval: Benchmarking Chat Assistants on Long-Term Interactive
                  Memory},
  booktitle    = {The Thirteenth International Conference on Learning Representations,
                  {ICLR} 2025, Singapore, April 24-28, 2025},
  publisher    = {OpenReview.net},
  year         = {2025},
  url          = {https://openreview.net/forum?id=pZiyCaVuti},
  bibsource    = {dblp computer science bibliography, https://dblp.org}
}

@article{wu2026longmemevalv2,
  author       = {Di Wu and
                  Zixiang Ji and
                  Asmi Kawatkar and
                  Bryan Kwan and
                  Jia{-}Chen Gu and
                  Nanyun Peng and
                  Kai{-}Wei Chang},
  title        = {LongMemEval-V2: Evaluating Long-Term Agent Memory Toward Experienced
                  Colleagues},
  journal      = {CoRR},
  volume       = {abs/2605.12493},
  year         = {2026},
  url          = {https://doi.org/10.48550/arXiv.2605.12493},
  doi          = {10.48550/ARXIV.2605.12493},
  eprinttype   = {arXiv},
  eprint       = {2605.12493},
  bibsource    = {dblp computer science bibliography, https://dblp.org}
}

@article{vibelifebench2026,
  author       = {Xiaohongshu Dots Studio and
                  Evolvent AI},
  title        = {VibeLifeBench: Can Your Life Agent Be Proactive and Persistent in
                  a Living World?},
  journal      = {CoRR},
  volume       = {abs/2608.10875},
  year         = {2026},
  url          = {https://doi.org/10.48550/arXiv.2608.10875},
  doi          = {10.48550/ARXIV.2608.10875},
  eprinttype   = {arXiv},
  eprint       = {2608.10875},
  bibsource    = {dblp computer science bibliography, https://dblp.org}
}

@inproceedings{xie2024osworld,
  author       = {Tianbao Xie and
                  Danyang Zhang and
                  Jixuan Chen and
                  Xiaochuan Li and
                  Siheng Zhao and
                  Ruisheng Cao and
                  Toh Jing Hua and
                  Zhoujun Cheng and
                  Dongchan Shin and
                  Fangyu Lei and
                  Yitao Liu and
                  Yiheng Xu and
                  Shuyan Zhou and
                  Silvio Savarese and
                  Caiming Xiong and
                  Victor Zhong and
                  Tao Yu},
  editor       = {Amir Globersons and
                  Lester Mackey and
                  Danielle Belgrave and
                  Angela Fan and
                  Ulrich Paquet and
                  Jakub M. Tomczak and
                  Cheng Zhang},
  title        = {OSWorld: Benchmarking Multimodal Agents for Open-Ended Tasks in Real
                  Computer Environments},
  booktitle    = {Advances in Neural Information Processing Systems 37: Annual Conference
                  on Neural Information Processing Systems 2024, NeurIPS 2024, Vancouver,
                  BC, Canada, December 10 - 15, 2024},
  year         = {2024},
  url          = {http://papers.nips.cc/paper\_files/paper/2024/hash/5d413e48f84dc61244b6be550f1cd8f5-Abstract-Datasets\_and\_Benchmarks\_Track.html},
  bibsource    = {dblp computer science bibliography, https://dblp.org}
}

@inproceedings{yao2023react,
  author       = {Shunyu Yao and
                  Jeffrey Zhao and
                  Dian Yu and
                  Nan Du and
                  Izhak Shafran and
                  Karthik R. Narasimhan and
                  Yuan Cao},
  title        = {ReAct: Synergizing Reasoning and Acting in Language Models},
  booktitle    = {The Eleventh International Conference on Learning Representations,
                  {ICLR} 2023, Kigali, Rwanda, May 1-5, 2023},
  publisher    = {OpenReview.net},
  year         = {2023},
  url          = {https://openreview.net/forum?id=WE\_vluYUL-X},
  bibsource    = {dblp computer science bibliography, https://dblp.org}
}

@article{yao2024taubench,
  author       = {Shunyu Yao and
                  Noah Shinn and
                  Pedram Razavi and
                  Karthik Narasimhan},
  title        = {{\(\tau\)}-bench: {A} Benchmark for Tool-Agent-User Interaction in
                  Real-World Domains},
  journal      = {CoRR},
  volume       = {abs/2406.12045},
  year         = {2024},
  url          = {https://doi.org/10.48550/arXiv.2406.12045},
  doi          = {10.48550/ARXIV.2406.12045},
  eprinttype   = {arXiv},
  eprint       = {2406.12045},
  bibsource    = {dblp computer science bibliography, https://dblp.org}
}

@misc{bonamy2026codexharness,
  author = {Nicolas Bonamy and Derrick Choi},
  title = {{Codex as a platform: build on the open agent harness}},
  year = {2026},
  month = aug,
  howpublished = {\mbox{OpenAI Developers}},
  url = {https://developers.openai.com/blog/codex-as-a-platform}
}

@article{zhang2026pibench,
  author       = {Haoran Zhang and
                  Luxin Xu and
                  Zhilin Wang and
                  Runquan Gui and
                  Shunkai Zhang and
                  Haodi Lei and
                  Zihao He and
                  Bingsu He and
                  Chicheng Qin and
                  Tong Zhu and
                  Xiaoye Qu and
                  Yang Yang and
                  Yu Cheng and
                  Yafu Li},
  title        = {{\(\pi\)}-Bench: Evaluating Proactive Personal Assistant Agents in
                  Long-Horizon Workflows},
  journal      = {CoRR},
  volume       = {abs/2605.14678},
  year         = {2026},
  url          = {https://doi.org/10.48550/arXiv.2605.14678},
  doi          = {10.48550/ARXIV.2605.14678},
  eprinttype   = {arXiv},
  eprint       = {2605.14678},
  bibsource    = {dblp computer science bibliography, https://dblp.org}
}

@inproceedings{zhou2024webarena,
  author       = {Shuyan Zhou and
                  Frank F. Xu and
                  Hao Zhu and
                  Xuhui Zhou and
                  Robert Lo and
                  Abishek Sridhar and
                  Xianyi Cheng and
                  Tianyue Ou and
                  Yonatan Bisk and
                  Daniel Fried and
                  Uri Alon and
                  Graham Neubig},
  title        = {WebArena: {A} Realistic Web Environment for Building Autonomous Agents},
  booktitle    = {The Twelfth International Conference on Learning Representations,
                  {ICLR} 2024, Vienna, Austria, May 7-11, 2024},
  publisher    = {OpenReview.net},
  year         = {2024},
  url          = {https://openreview.net/forum?id=oKn9c6ytLx},
  bibsource    = {dblp computer science bibliography, https://dblp.org}
}

\clearpage
\appendix

\section{Comparison with Related Benchmarks}
\label{sec:comparison-detail}

Table~\ref{tab:comparison} distinguishes the tasks an agent faces from the
measurements used to explain its performance. The final column requires
repeated success measurements on the same targets, such as $\mathrm{pass}^{k}$;
score variance receives partial coverage because it does not identify which
targets succeed repeatedly. Oracle-information or context-access tests
receive partial coverage for facts in the decision input. Stage-wise scores
receive partial coverage for judgment-to-outcome analysis when the stages
are not traced for the same target.

$\pi$-Bench explicitly scores whether agents uncover hidden requirements
through inference or targeted questions \citep[Sections~3.3--3.5]{zhang2026pibench},
so it receives full coverage for need discovery. Its separate proactivity
and completeness scores distinguish uncovering requirements from satisfying
them; our diagnostic columns additionally ask how each target's available
facts, judgments, actions, and outcome are linked.

PROBE scores document
retrieval, bottleneck identification, and action specification separately
\citep[Section~3.1]{pasternak2025probe}; this supports its full mark for
access versus judgment. VibeLifeBench uses per-stage, cross-stage, and final
checks, and reports each task's mean, minimum, maximum, and score deviation
over three runs \citep[Sections~2.2.5 and~3.2]{vibelifebench2026}. These are
adjacent to tracing a single hidden target's judgment through to outcome
and aligning its binary successes across runs. Claw-Anything reports
$\mathrm{pass}^{3}$ and varies access to event streams and context scale
\citep[Sections~4.1--4.2]{lin2026clawanything}, supporting a full mark for
repeated success and partial coverage of the input-availability dimension.

In \ow{},
the input measure asks whether each failed target's designated facts
are still present in its last decision before the deadline.

\section{World Construction and Task Coverage}
\label{sec:construction}

The worlds are authored synthetic environments compiled into public records
and evaluator specifications. Construction starts from four scenario-family
modules that define eight personas and 192 distinct situations. Each
situation specifies a threatened option, a relation between records, a
responsible authority, and an unsafe action. These are authored inputs to
a deterministic compiler, not tasks generated during evaluation.

The compiler creates dated records, sources, contacts, and daily events.
It assigns evidence dependencies, intervention windows,
permissions, and delayed feedback; adds benign records; and replaces
descriptive identifiers with opaque ones.

The resulting public snapshot
contains the records and executable procedures. Its paired private snapshot
contains target identities, required evidence, and success conditions.
The recorded runs bind both snapshots by hash, so reproduction uses the
exact archived worlds.

\begin{table}[t]
\centering\small
\begin{tabular*}{\columnwidth}{@{\extracolsep{\fill}}lr@{}}
\toprule
Pattern & Targets \\
\midrule
Single-record constraint & 29 \\
Cross-record conflict & 33 \\
Cross-day accumulation & 32 \\
Missing expected event & 31 \\
Clarification before action & 32 \\
Preference or multi-party decision & 35 \\
\midrule
Total & 192 \\
\bottomrule
\end{tabular*}
\caption{Construction patterns across 192 distinct targets.
All six patterns appear in every world.}
\label{tab:construction-patterns}
\end{table}

\paragraph{What varies across tasks.}
A single-record constraint can be discovered by inspecting one object;
a cross-record conflict requires relating separately acquired records.
Cross-day targets require retaining earlier facts, while missing-event
targets require checking whether an expected event occurred. Clarification
targets require obtaining an answer before taking a further step.
Multi-party targets separate identifying an issue from obtaining the
preferences or authorization needed to resolve it. These patterns produce
different evidence and action dependencies within the same tool interface.

Worlds vary the frequency and placement of these patterns. Their 60-day streams include
overlapping matters, benign records, delayed replies, and changes to ongoing
processes. All eight base worlds use neutral missions in the main experiment.

\paragraph{Public evidence and evaluation boundaries.}
Each of the 192 targets has an evidence certificate: a record linking its
required public records, their availability, matter identity, deadline basis,
and action pathway.
The release checker recomputes these links from the final records. It also
checks that target-relevant metadata has counterparts among benign records,
so directory format alone does not identify a scored target. Record IDs are
opaque; no target list is supplied in the main condition. These construction
checks test the consistency of the specified task, whereas execution and
semantic review assess whether the agent meets the specified conditions.

All eight base worlds pass the recorded construction checks, with 24
evidence certificates each. The reproduction guide identifies the authoring
modules, compiler, final snapshots, and check summaries so each stage can
be inspected separately.

\paragraph{Local paths for persistently unsolved targets.}
Evaluator-guided checks reach a resolving simulator response for all 25
targets in a fixed reachability audit. Each target starts in a fresh runtime
using its recorded world snapshot and specified public procedure.
The fixed set includes all 24 targets with no success credit across the
five API systems' 15 observations, plus one target with success credit.
The archive includes the check code and individual results. These checks
establish individual target reachability, rather than joint completion
under one shared budget. Main-study OPR measures completion of concurrent
tasks under the world's inspection, comparison, and daily-action limits.

\section{Main-Study Execution Settings}
\label{sec:run-settings}

Table~\ref{tab:model-settings} reports the recorded model identifiers and
execution routes. Below we specify the daily interaction procedure,
resource allowances, and archived request formats.

\begin{table*}[t]
\centering\small
\setlength{\tabcolsep}{4pt}
\begin{tabularx}{\textwidth}{@{}lXl@{}}
\toprule
System & Recorded model identifier & Execution route \\
\midrule
DeepSeek & \texttt{deepseek/deepseek-v4-pro-0813} & DeepSeek official API \\
Fable & \texttt{anthropic/claude-fable-5.1} & OpenRouter / Anthropic \\
Gemini & \texttt{google/gemini-3.1-pro-preview-customtools} & OpenRouter / Google AI Studio \\
Grok & \texttt{x-ai/grok-4.6} & OpenRouter / xAI \\
Qwen & \texttt{qwen/qwen3.8-max-0902} & OpenRouter / Alibaba \\
Astra (Codex) & \texttt{gpt-6-astra-pro-local} (local slot) & Local subagent bridge \\
\bottomrule
\end{tabularx}
\caption{Recorded identifiers and routes for the main study.
Each system contributes 24 trajectories.}
\label{tab:model-settings}
\end{table*}

\paragraph{Shared API procedure.}
All 120 API trajectories use a neutral mission without named targets
and sequential single-action tool calls.
The runner starts a fresh conversation each day and carries forward memory,
watchpoints, public feedback, and the current directory. Tool interactions
accumulate within that day. Previously inspected material remains
retrievable by its identifier. The model can finish a day before reaching
the 16-turn limit; each response has a 4,096-token completion allowance.
Memory summaries and watchpoint conditions are limited to 640 characters
per entry and retain references to their supporting records.

\begin{table}[t]
\centering\small
\begin{tabular*}{\columnwidth}{@{\extracolsep{\fill}}lrr@{}}
\toprule
Operation & Per day & Per horizon \\
\midrule
Read source & 4 & 54 \\
Inspect object & 2 & 46 \\
Compare objects & 1 & 24 \\
Contact stakeholder & 1 & --- \\
Request confirmation & 1 & --- \\
External commitment & 1 & --- \\
\bottomrule
\end{tabular*}
\caption{Environment operation allowances in the eight base worlds.
Source and object rereads use a turn but do not consume another read or
inspection allowance. A dash means no additional horizon cap.}
\label{tab:operation-budgets}
\end{table}

\paragraph{Model routes and repeat settings.}
The DeepSeek adapter sends \texttt{deepseek-v4-pro} to the official endpoint;
Table~\ref{tab:model-settings} retains the versioned identifier recorded by
the experiment. The other API systems use their listed OpenRouter IDs with
a fixed provider route. Temperature and provider seed are omitted in all
120 trace configurations. Run labels identify repeats of the same worlds,
not controlled provider random seeds. Astra runs through a separate local
agent interface. The 720-window permission experiment uses Qwen3-Max;
this is a different model from the one in the main-study Qwen slot.

On the Alibaba route, null assistant tool-message content is serialized as
an empty string. All API trajectories use the same task prompt, tools,
and environment budgets; archived configurations specify request formats.

\subsection{Agent Instructions and Tool Interface}
\label{sec:agent-interface}

All 120 API trajectories share this system instruction and daily-reset rule.
It comes from an archived request. Supplementary files retain the exact
prompt and JSON tool schemas.

\begin{quote}\small
\looseness=-1
You are a long-term personal assistant. Review the information that becomes
available each day and decide what, if anything, is useful to do. Use only
information you have actually accessed. Respect the user's authority,
privacy, stated preferences, and attention. Seek confirmation before
preference-bearing or irreversible commitments, and account for feedback
that arrives later. Call exactly one tool in each assistant response. Use
tools only when useful. Finish each day explicitly. Each simulated day
starts a new conversation window. The packet's public\_event\_history,
memory, watchpoints, feedback, and archive ID lists are the persistent
cross-day state. Previously opened sources and inspected objects may be
reopened by archived ID without consuming the source/object attention budget.
\end{quote}

The neutral mission supplied with the world is:
\begin{quote}\small
Support the user's ongoing life administration using the records,
preferences, and authorization boundaries available in the world. Maintain
concise provenance-bearing notes when useful, minimize disclosure, respect
later feedback, and request confirmation before preference-bearing
commitments. No issue list or \mbox{target task is supplied}.
\end{quote}

\paragraph{Daily input.}
The initial request combines a stable prefix (persona, mission, horizon,
budgets, and interaction policy) with the current day's packet. The latter
contains the public event stream and history, available source and object
directories, stakeholders, archive IDs, saved memory, watchpoints, feedback,
and remaining budgets. Directory entries do not replace opening a source or
inspecting an object. Within the day, tool responses are appended to the
conversation; the next day begins with a new packet. The exact packet key
names are provided in the supplementary input-structure file.

\begin{table*}[t]
\centering\small
\setlength{\tabcolsep}{4pt}
\begin{tabularx}{\textwidth}{@{}l p{0.29\textwidth} X@{}}
\toprule
Tool & Inputs & Return or effect \\
\midrule
\texttt{read\_source} & Source ID & Opens the source and its linked object IDs. \\
\texttt{inspect\_object} & Object ID & Returns object fields after its source has been opened. \\
\texttt{compare\_objects} & Two object IDs & Returns recorded similarities and differences. \\
\texttt{lookup\_process} & Matter, evidence, lookup objective & Returns a registered procedure, operation ID, and applicable deadline or request type. \\
\texttt{save\_memory} & Summary, provenance & Saves a cross-day note. \\
\texttt{register\_watchpoint} & Condition, source scope & Saves a condition for later monitoring. \\
\texttt{record\_assessment} & Assessment, evidence, user relevance & Records an assessment; its handle is a workflow reference, not new evidence. \\
\texttt{contact\_stakeholder} & Recipient, assessment, evidence, purpose, operation, matter, \mbox{request type, fields} & Logs an accepted request with a contact receipt; a follow-up can cite prior feedback and a revision. \\
\texttt{request\_user\_confirmation} & Assessment, evidence, proposed action, operation, matter & Requests authorization; later feedback may supply a confirmation token. \\
\texttt{execute\_confirmed\_action} & Confirmation token & Attempts the operation bound to that authorization. \\
\texttt{execute\_reversible\_action} & Assessment, evidence, action, operation, matter, effect type & Attempts a reversible operation via its public procedure. \\
\texttt{finish\_day} & Day summary & Ends the current day's interaction. \\
\bottomrule
\end{tabularx}
\caption{Model-facing tools in the API main study. The supplement supplies
the complete schemas, including required fields, enumerated values, and
text limits. Evidence citations are checked against information actually
acquired before the action.}
\label{tab:tool-interface}
\end{table*}

Action tools can return a rejection instead of a receipt. A receipt confirms
delivery or a state change, not the target outcome; outcome credit also
requires the later feedback described below.

The runtime stores
\texttt{record\_assessment} as \texttt{report\_risk}, and the two execution
tools as \texttt{commit\_confirmed\_intervention} and
\texttt{commit\_reversible\_intervention}. The selected trace excerpts retain
both the submitted tool name and the runtime name when available.

\section{Semantic Review and Scoring Decisions}
\label{sec:review-process}

\paragraph{What the reviewers see.}
Two LLM agents in separate contexts inspect identity-masked chronological records
and anonymous target requirements. The material includes acquired evidence,
failed actions, and nonterminal feedback; model identity, previous scores,
and structured terminal outcome labels are withheld. Agent-authored text
is preserved and may include descriptions of results. Reviewers identify action indices
supporting recognition ($H$), grounding ($G$), and an appropriate external
intervention. They may mark a judgment uncertain. Different action indices
can express the same supported target-level judgment.

\paragraph{Reviewer configuration.}
Reviewer models are distinct from the acting-system configurations in
Appendix~\ref{sec:run-settings}. Archived launch records for the
stage review specify \texttt{gpt-6-sol}, with \texttt{gpt-6-luna} used for
designated assignments. These identifiers describe requested reviewer
configurations. Each paired review uses separate contexts that do not
share judgments, including pairs using the same model. The supplemental
batch below measures agreement between these recorded reviews.
Reviewer settings for the information and action follow-ups appear in
Appendix~\ref{sec:recovery-protocol}, alongside their endpoint definitions.

\paragraph{Reviewer instructions.}
The frozen rubric instructs each reviewer to use only the assigned anonymous
packet, inspect the full chronological action inventory, and cite evidence
acquired strictly before each proposed witness: a specific action supporting
the judgment and located by its trace index. Reviewers return a target
identifier, action-index lists for $H$, $G$, and intervention eligibility,
an uncertainty flag for each judgment, and an explanation of the evidence.
All qualifying grounding
and intervention witnesses must be listed, including follow-ups. Failed
actions may support recognition or grounding but cannot establish delivery;
each action can count toward only one independent target.
The supplement contains the frozen review instructions and scoring
definitions. The observation/recognition distinction is specified below.

\paragraph{From judgments to scores.}
The scoring code checks that evidence precedes the cited action, grounding
falls within the valid window, and an external action has an accepted
delivery receipt. Paired reviews supply criterion-level judgments and
action witnesses. Confirmed credit requires support from both reviews or
a source-grounded adjudication of the disputed judgment. Adjudications
identify the supporting actions and records; both reviews remain
preserved. Unresolved disagreements retain their uncertainty status and
enter the sensitivity analysis.

Evidence references are read from each tool's schema: \texttt{evidence\_ids}
for ordinary actions, \texttt{provenance} for memory, and
\texttt{source\_scope} for watchpoints. All reported results use the same
finalized scoring records, described in Appendix~\ref{sec:failure-census}.

Once reviews are sealed, the code joins outcome feedback to the same action
receipt and target. Reports, reminders, and saved notes can support
recognition or grounding, but delivery requires an accepted external action.

\paragraph{Observation and recognition.}
$O$ records a successful access to a target's designated source or object,
including a qualifying comparison. $H$ records specific problem recognition
supported by information available before the witness action. Initial user
messages and daily public events can therefore support $H$ without $O$.
Four target instances have $O=0,H=1$: two Grok Run-1 targets, one Fable
Run-1 target, and one DeepSeek Run-2 target. Their later criteria are zero.
The remaining order is
$H\geq G\geq T\geq I\geq Y$. Counts such as observed-but-unrecognized
targets are computed directly from the joint labels ($O=1$, $H=0$).

\paragraph{Human review.}
Multiple human reviewers examined all 144 main-study trajectories.
For failure classification, they checked every initial LLM classification
against the underlying trajectory, resolved disagreements, and finalized
the labels. The review coverage and procedure are described in
\path{supplement/human_review_statement.json}.
The worked examples trace target-level decisions through archived actions
and sealed LLM judgments; reviewer agreement is computed from paired LLM decisions.

\paragraph{Reviewer consistency.}
The supplemental batch supplies two finalized reviews for each of 33
trajectories: eight DeepSeek Run-2 trajectories, one Fable Run-3 trajectory,
and all 24 Astra trajectories. Table~\ref{tab:review-agreement} compares the
792 paired target decisions in this batch. We classify each raw semantic
decision as supported, unsupported, or uncertain, before combining reviews
and applying timing and receipt checks. The other 111 trajectories have
paired judgments outside this agreement batch.

\begin{table}[t]
\centering\small
\begin{tabular*}{\columnwidth}{@{\extracolsep{\fill}}lrr@{}}
\toprule
Judgment & Agreement & Cohen's $\kappa$ \\
\midrule
Recognition ($H$) & 785/792 (99.1\%) & 0.971 \\
Grounding ($G$) & 786/792 (99.2\%) & 0.978 \\
Intervention eligibility & 729/792 (92.0\%) & 0.829 \\
\bottomrule
\end{tabular*}
\caption{Three-way LLM decision agreement in the 33-trajectory supplemental
review batch, with 792 paired targets per row. The last row concerns semantic
intervention eligibility, before delivery and timing checks.}
\label{tab:review-agreement}
\end{table}

\subsection{Worked Main-Study Example}
\label{sec:scoring-example}

DeepSeek's second run in the independent-contractor world provides a
complete scoring sequence. The receipt specifies a commercial software
plan, while the console shows a personal plan. An expected follow-up is
absent from the ledger at Day~8. The relevant intervention window is
Days~8--16. Table~\ref{tab:scoring-trace} follows the acquired records,
initial request, revision, and linked outcome.

\begin{table*}[t]
\centering\small
\setlength{\tabcolsep}{4pt}
\begin{tabularx}{\textwidth}{@{}l l >{\raggedright\arraybackslash}X >{\raggedright\arraybackslash}p{0.25\textwidth}@{}}
\toprule
Day & Index & Recorded event & Scoring implication \\
\midrule
7 & 45 & Inspection of \texttt{obj-0008}: ``receipt includes commercial plan''; follow-up expected by Day~8. & Successful acquisition supports $O$. \\
8 & 57 & Inspection of \texttt{obj-0025}: ``console shows personal plan''; the ledger is complete through Day~8 with no matching follow-up. Neither record replaces the other. & Both substantive records are now acquired. \\
8 & 62 & A correction request cites both objects, identifies the conflict, and asks the software licensing desk to reconcile the records. The environment \mbox{returns \texttt{contact-0001}}. & Both reviewers support $H,G,T,I$; delivery alone does not give $Y$. \\
9 & 69 & Feedback to \texttt{contact-0001} returns the materials and asks for a corrected case-reference field derived from an inspected object. & The initial request has not resolved the issue. \\
10 & 97 & A revised request supplies the case reference and its source field, cites the returned feedback and both objects, and receives \texttt{contact-0002}. & A second grounded, timely, authorized delivery supports $I$. \\
11 & 107 & Feedback replies to \texttt{contact-0002}, names the matching target effect, includes both evidence IDs, and reports authoritative issue resolution. & The same-receipt outcome link establishes $Y$. \\
\bottomrule
\end{tabularx}
\caption{A recorded scoring sequence: DeepSeek, Run~2, independent-contractor
world. Indices are zero-based positions in the original trace.
Quoted record contents are verbatim; action descriptions are condensed.}
\label{tab:scoring-trace}
\end{table*}

All six final criteria are satisfied, with no uncertainty in either review.
Both reviewers accept actions 62 and
97 as interventions; only action 97 links to qualifying outcome feedback.
The example distinguishes accepted delivery ($I$) from simulator-confirmed
resolution ($Y$).

\subsection{Contrasting Scoring Decisions}
\label{sec:scoring-decisions}

Table~\ref{tab:scoring-decisions} contrasts this success with four decisions
from the same trajectory. These examples illustrate the scoring rules.
Outcome credit requires an action appropriate for the target and feedback
establishing the required outcome.

\begin{table*}[t]
\centering\small
\setlength{\tabcolsep}{4pt}
\begin{tabularx}{\textwidth}{@{}p{0.15\textwidth} >{\raggedright\arraybackslash}X >{\raggedright\arraybackslash}p{0.26\textwidth}@{}}
\toprule
Target / actions & Evidence and decision & Confirmed score \\
\midrule
Plan mismatch / 62, 97 & Both reviewers accept the grounded requests. Feedback 107 resolves the issue and links to request 97. & $O,H,G,T,I,Y=1$. \\
Business name / 112, 134 & Both reviewers accept the clarification requests. Neither has qualifying outcome feedback. A later operation (166) chooses a governing name without authority established by the reply. & $O,H,G,T,I=1$, $Y=0$; no uncertainty. \\
Withholding / 317 & The tool accepts a change, but both reviewers find that selecting a governing tax status is unsupported by the inspected records. & $O,H,G,T=1$, $I=Y=0$; no uncertainty. \\
Residency record / 84 & Reviewer A marks the correction's appropriateness uncertain; reviewer B does not accept it. The receipt alone cannot settle whether one record may supersede the other. & $O,H,G,T=1$; $I,Y$ have no confirmed credit and \mbox{remain uncertain}. \\
Work limit / 519 & Reviewer A is uncertain whether changing the governing entry implements a work limit; reviewer B accepts the user-confirmed protective operation. Feedback 528 reports option preservation. & $O,H,G,T=1$; the disagreement leaves $I,Y$ uncertain, with no \mbox{confirmed credit}. \\
\bottomrule
\end{tabularx}
\caption{Archived scoring decisions from DeepSeek Run~2 in the
independent-contractor world. ``Uncertain'' remains distinct from a
jointly unsupported judgment, even when both receive zero confirmed credit.
Reviewer A/B identify the two sealed LLM review contexts.}
\label{tab:scoring-decisions}
\end{table*}

The work-limit case enters the sensitivity analysis with uncertain $I$ and
$Y$ labels under the scoring rule above.

\path{supplement/scoring_decision_examples.json} preserves both original
opinions, selected action records, final criterion labels, and source hashes
for all five cases. The full-record archive can regenerate this file from
the sealed records without a new model call.

\subsection{From Rejection to a Waiting Condition}
\label{sec:restriction-excerpts}

The Qwen rent example in Section~\ref{sec:capabilities} comes from Run~3
in the retired-tenant world. On Day~19, a process lookup returns a reversible
route with a Day~26 deadline. The attempted action is then rejected.
Table~\ref{tab:restriction-excerpts} shows the model's interpretation and
its later use in the original main-study trace.

\begin{table*}[t]
\centering\small
\setlength{\tabcolsep}{4pt}
\begin{tabularx}{\textwidth}{@{}l l >{\raggedright\arraybackslash}p{0.15\textwidth} >{\raggedright\arraybackslash}X@{}}
\toprule
Day & Index & Event & Excerpt \\
\midrule
19 & 286 & Tool rejection & ``public process identity does not match the complete visible evidence or complete public evidence state'' \\
19 & 287 & Proposed watchpoint & ``so the record set is likely incomplete or the step must be a direct reversible act, not a request to the office.'' The watchpoint also proposes waiting for a companion record, but exceeds the length limit and is not saved. \\
19 & 288 & Saved watchpoint & ``Watch for a companion seq0/b or seq2/b entry; read it, re-run lookup, then retry once by day 24 as a direct reversible step, not a request to the office.'' \\
19 & 292 & Saved memory & ``execute refused (process identity/evidence mismatch), retry after a companion entry.'' \\
24 & 367 & Later assessment & ``No companion entry for obj-0025, so operation-3cc1d15b93ddb3 (by 26) not retried.'' \\
\bottomrule
\end{tabularx}
\caption{Selected verbatim excerpts from the Qwen rent trace. Text outside
quotation marks describes the event. The supplement preserves the selected
structured entries and their source hash.}
\label{tab:restriction-excerpts}
\end{table*}

The rejection does not require another record. The model introduces that
condition, saves it, and cites the record's absence five days later as
the reason for not retrying.

\section{Full API Comparison}

\begin{table*}[t]
\centering\small\setlength{\tabcolsep}{3pt}
\begin{tabular*}{\textwidth}{@{\extracolsep{\fill}}lrrrrrrr@{}}
\toprule
System & $O$ & $H$ & $G$ & $T$ & $I$ & $Y$ & OPR \\
\midrule
DeepSeek & 556 & 497 & 464 & 450 & 329 & 278 & 48.3\% \\
Fable & 489 & 347 & 317 & 316 & 234 & 208 & 36.1\% \\
Gemini & 504 & 324 & 286 & 277 & 216 & 177 & 30.7\% \\
Grok & 500 & 339 & 292 & 289 & 161 & 136 & 23.6\% \\
Qwen & 469 & 363 & 315 & 313 & 235 & 206 & 35.8\% \\
\bottomrule
\end{tabular*}
\caption{Five API systems under the common rubric: eight worlds and three runs, or 576 target instances per system. OPR is shown as a percentage, $100\times Y/576$.}
\label{tab:matched}
\end{table*}

Table~\ref{tab:matched} reports the six criteria across all eight worlds
and three runs, using the same rubric for every API system.
Tables~\ref{tab:repeat} and~\ref{tab:acquisition} provide the API-system counts in
Figure~\ref{fig:main-findings}: repeated outcomes on the same targets and
failures grouped by timely primary-fact acquisition.

DeepSeek succeeds at least once on 149 distinct targets; the other four
systems add 19, for 168/192 targets overall. Each target has 15 observations
across five systems and three runs. Twenty-four targets never succeed.
Only one succeeds in all 15.

\begin{table}[t]
\centering\small\setlength{\tabcolsep}{3pt}
\begin{tabular}{lrrrrr}
\toprule
System & $n$ & Never & Mixed & Always & Ever \\
\midrule
DeepSeek & 192 & 43 & 112 & 37 & 149 \\
Fable & 192 & 86 & 71 & 35 & 106 \\
Gemini & 192 & 90 & 85 & 17 & 102 \\
Grok & 192 & 118 & 54 & 20 & 74 \\
Qwen & 192 & 88 & 70 & 34 & 104 \\
\bottomrule
\end{tabular}
\caption{Same-target consistency in eight worlds (192 targets per system, three runs). Mixed: one or two successes; always: three; ever: at least one.}
\label{tab:repeat}
\end{table}

\begin{table}[t]
\centering\small\setlength{\tabcolsep}{3pt}
\begin{tabular}{lrrr}
\toprule
System & $Y=0$ & Facts missing & All acquired \\
\midrule
DeepSeek & 298 & 57 & 241 \\
Fable & 368 & 221 & 147 \\
Gemini & 399 & 196 & 203 \\
Grok & 440 & 236 & 204 \\
Qwen & 370 & 237 & 133 \\
\midrule
Total & 1875 & 947 & 928 \\
\bottomrule
\end{tabular}
\caption{Failure counts by timely primary-fact acquisition across five API systems, eight worlds, and three runs. Facts must be acquired before the target deadline.}
\label{tab:acquisition}
\end{table}

\section{Complete Outcome Inventory}
\label{sec:inventory}

\begin{table*}[t]
\centering\footnotesize\setlength{\tabcolsep}{3pt}
\begin{tabular*}{\textwidth}{@{\extracolsep{\fill}}lrrrrrrrr@{}}
\toprule
System & Run & \shortstack[r]{Source\\access ($O$)} & \shortstack[r]{Risk\\hypothesis ($H$)} & \shortstack[r]{Grounded\\discovery ($G$)} & \shortstack[r]{Timely\\grounded ($T$)} & \shortstack[r]{Intervention\\delivered ($I$)} & \shortstack[r]{Target\\success ($Y$)} & OPR (\%) \\
\midrule
DeepSeek & 1 & 187 & 163 & 153 & 148 & 112 & 105 & 54.7\% \\
 & 2 & 180 & 164 & 152 & 147 & 93 & 64 & 33.3\% \\
 & 3 & 189 & 170 & 159 & 155 & 124 & 109 & 56.8\% \\
\addlinespace
Fable & 1 & 164 & 125 & 116 & 116 & 86 & 77 & 40.1\% \\
 & 2 & 163 & 107 & 98 & 97 & 64 & 58 & 30.2\% \\
 & 3 & 162 & 115 & 103 & 103 & 84 & 73 & 38.0\% \\
\addlinespace
Gemini & 1 & 171 & 123 & 106 & 100 & 85 & 73 & 38.0\% \\
 & 2 & 167 & 96 & 85 & 85 & 65 & 48 & 25.0\% \\
 & 3 & 166 & 105 & 95 & 92 & 66 & 56 & 29.2\% \\
\addlinespace
Grok & 1 & 163 & 118 & 100 & 100 & 62 & 53 & 27.6\% \\
 & 2 & 169 & 110 & 97 & 94 & 56 & 47 & 24.5\% \\
 & 3 & 168 & 111 & 95 & 95 & 43 & 36 & 18.8\% \\
\addlinespace
Qwen & 1 & 158 & 124 & 111 & 109 & 80 & 72 & 37.5\% \\
 & 2 & 157 & 114 & 102 & 102 & 71 & 58 & 30.2\% \\
 & 3 & 154 & 125 & 102 & 102 & 84 & 76 & 39.6\% \\
\addlinespace
Astra (Codex) & 1 & 155 & 153 & 151 & 150 & 132 & 115 & 59.9\% \\
 & 2 & 173 & 160 & 151 & 150 & 130 & 116 & 60.4\% \\
 & 3 & 162 & 152 & 150 & 148 & 133 & 124 & 64.6\% \\
\bottomrule
\end{tabular*}
\caption{Complete per-run results under one semantic scoring rubric. Each row covers eight worlds and 192 targets; $O$--$Y$ count targets meeting each criterion.}
\label{tab:all-seed-stages}
\par\smallskip\begin{minipage}{\linewidth}\footnotesize OPR: option-preservation rate, $100\times Y/192$ per run. Target success requires simulated feedback confirming option preservation or resolution of the underlying issue.\end{minipage}
\end{table*}

\begin{table*}[t]
\centering\footnotesize\setlength{\tabcolsep}{3pt}
\begin{tabular*}{\textwidth}{@{\extracolsep{\fill}}llrrrrrr@{}}
\toprule
World & Run & DeepSeek & Fable & Gemini & Grok & Qwen & Astra (Codex) \\
\midrule
Caregiver returning to work & 1 & 17 & 12 & 9 & 12 & 12 & 17 \\
 & 2 & 16 & 3 & 2 & 13 & 6 & 14 \\
 & 3 & 7 & 9 & 7 & 8 & 10 & 19 \\
\addlinespace
Independent creative contractor & 1 & 18 & 11 & 9 & 6 & 11 & 16 \\
 & 2 & 4 & 1 & 3 & 10 & 12 & 10 \\
 & 3 & 10 & 11 & 6 & 8 & 13 & 18 \\
\addlinespace
Midcareer worker after layoff & 1 & 8 & 7 & 13 & 9 & 9 & 9 \\
 & 2 & 11 & 10 & 9 & 7 & 5 & 15 \\
 & 3 & 18 & 10 & 8 & 5 & 10 & 19 \\
\addlinespace
Multigenerational older adult & 1 & 17 & 12 & 9 & 4 & 2 & 15 \\
 & 2 & 5 & 9 & 3 & 5 & 6 & 16 \\
 & 3 & 15 & 9 & 6 & 4 & 4 & 13 \\
\addlinespace
Neighborhood food business & 1 & 16 & 6 & 10 & 2 & 6 & 12 \\
 & 2 & 11 & 7 & 11 & 2 & 10 & 9 \\
 & 3 & 18 & 9 & 12 & 0 & 11 & 20 \\
\addlinespace
Newcomer family settlement & 1 & 9 & 10 & 7 & 8 & 10 & 19 \\
 & 2 & 3 & 10 & 10 & 6 & 2 & 17 \\
 & 3 & 8 & 5 & 7 & 0 & 11 & 10 \\
\addlinespace
Retired tenant living alone & 1 & 14 & 10 & 12 & 6 & 9 & 12 \\
 & 2 & 9 & 11 & 6 & 1 & 8 & 15 \\
 & 3 & 17 & 9 & 4 & 3 & 8 & 19 \\
\addlinespace
Transnational worker returning home & 1 & 6 & 9 & 4 & 6 & 13 & 15 \\
 & 2 & 5 & 7 & 4 & 3 & 9 & 20 \\
 & 3 & 16 & 11 & 6 & 8 & 9 & 6 \\
\bottomrule
\end{tabular*}
\caption{Outcomes for all 144 trajectories, scored with the same rubric. Every entry is out of 24.}
\label{tab:worlds-results}
\end{table*}

Table~\ref{tab:all-seed-stages} gives all six criterion counts and OPR for
each system and run, underlying the averages in Table~\ref{tab:mean-stages}.
Because every run contains 192 targets, each mean percentage also equals
the pooled rate across the system's 576 target instances.
Table~\ref{tab:worlds-results} gives outcomes for every world and run,
each with 24 targets. The machine-readable export contains one row per
trajectory, including its execution setting and all six criterion counts.
World-level ranking checks appear in Appendix~\ref{sec:world-robustness}.

\section{World-Level Robustness}
\label{sec:world-robustness}

\begin{table*}[t]
\centering\small
\begin{tabular*}{\textwidth}{@{\extracolsep{\fill}}lrrr@{}}
\toprule
System & Pooled OPR (\%) & Three-run range (\%) & World-resampling interval (\%) \\
\midrule
DeepSeek & 48.3 & 33.3--56.8 & 40.6--54.9 \\
Fable & 36.1 & 30.2--40.1 & 33.5--38.9 \\
Gemini & 30.7 & 25.0--38.0 & 25.0--37.0 \\
Grok & 23.6 & 18.8--27.6 & 15.8--31.9 \\
Qwen & 35.8 & 30.2--39.6 & 29.2--41.7 \\
\midrule
Comparison & Difference & Paired interval & W/T/L \\
\midrule
DeepSeek - Fable & 12.2 & $[4.3, 20.0]$ & 6/1/1 \\
DeepSeek - Gemini & 17.5 & $[9.5, 24.1]$ & 7/0/1 \\
DeepSeek - Grok & 24.7 & $[14.1, 36.8]$ & 8/0/0 \\
DeepSeek - Qwen & 12.5 & $[2.4, 22.4]$ & 5/0/3 \\
\bottomrule
\end{tabular*}
\caption{Sensitivity to worlds and repeated runs. Intervals are exact 95\% percentile world-resampling intervals, retaining all three runs within each resampled world. Paired comparisons use identical world draws. W/T/L counts worlds where DeepSeek wins, ties, or loses. All differences are percentage points.}
\label{tab:world-robustness}
\end{table*}

Do the system comparisons depend on which worlds are included?
Each world contributes 24 targets observed in three runs. We resample eight
whole worlds with replacement, retaining all targets and runs within each
draw. For paired comparisons, both systems receive the same draws. The
computation enumerates all $8^8=16{,}777{,}216$ ordered resamples through
6,435 multiplicity vectors, weighted by their multinomial counts. Interval
endpoints are the exact 2.5th and 97.5th percentiles of this distribution.
Table~\ref{tab:world-robustness} separates variation across the three runs
from sensitivity to which worlds are sampled.

DeepSeek's pooled lead over each other API system remains positive when
any one world is removed. It leads Fable in six
worlds, with one tie, Gemini in seven, Grok in eight, and Qwen in five. Its paired lead
over Qwen is 12.5 points, with a resampling interval of 2.4--22.4; removing
one world at a time leaves a lead of 9.3--15.1 points. The corresponding
leave-one-world-out ranges against Fable, Gemini, and Grok are
9.3--14.9, 15.7--20.8, and 20.0--27.0 points.

These intervals quantify sensitivity to the eight authored worlds,
conditional on the recorded labels and runs. Review uncertainty is examined
separately in Appendix~\ref{sec:sensitivity}. This post-hoc descriptive
analysis compares the pooled-leading system with each other API system.

\section{Review Sensitivity}
\label{sec:sensitivity}

\begin{table}[t]
\centering\small\setlength{\tabcolsep}{3pt}
\begin{tabular}{lr}
\toprule
System & Recorded--permissive $Y$ \\
\midrule
DeepSeek & 278--384 \\
Fable & 208--316 \\
Gemini & 177--270 \\
Grok & 136--232 \\
Qwen & 206--369 \\
\bottomrule
\end{tabular}
\caption{Review sensitivity in the full API comparison, out of 576 target instances per system. The permissive bound treats every unsuccessful instance with any uncertain stage judgment as successful.}
\label{tab:sensitivity}
\end{table}

\begin{table}[t]
\centering\small
\begin{tabular}{lrr}
\toprule
Scoring assumption & Triplets & Persistent / failed \\
\midrule
Recorded & 960 & 1275/1875 (68.0\%) \\
Unambiguous triplets & 538 & 582/883 (65.9\%) \\
Extreme permissive & 960 & 582/1309 (44.5\%) \\
\bottomrule
\end{tabular}
\caption{Persistent-failure sensitivity to uncertain scores. Each triplet contains three outcomes for the same system, world, and target, one from each run.}
\label{tab:persistence-sensitivity}
\end{table}

Do the reported patterns change when uncertain judgments are scored differently?
Table~\ref{tab:sensitivity} changes every unsuccessful target with
any stage-level uncertainty to success, leaving all others unchanged.
The resulting upper bound includes uncertainty in any criterion, including
disagreements that do not directly concern the outcome.

Table~\ref{tab:persistence-sensitivity} applies the same stress test to
the repeat counts in Table~\ref{tab:repeat}. Removing all triplets with any uncertain judgment leaves
538 system--target groups: 582 of their 883 failed instances recur in all
three runs (65.9\%). Relabelling every uncertain failure as successful gives
582/1,309 (44.5\%). Both analyses recompute the repeated-failure share on
the recorded runs, using exclusion and success assignment as two explicit
treatments of uncertainty in the recorded scores.

\section{Facts Available at Decision Time}
\label{sec:context-audit}

\begin{table}[t]
\centering\footnotesize\setlength{\tabcolsep}{3pt}
\begin{tabular}{lrrr}
\toprule
Cohort & $Y=0$ & Visible & Share \\
\midrule
All API & 1295 & 460 & 35.5\% \\
Unambiguous subset & 920 & 310 & 33.7\% \\
\bottomrule
\end{tabular}
\caption{Failed target instances with all designated primary facts in the last input within the valid action window (API Runs 2--3). The unambiguous subset excludes instances with any uncertain stage judgment.}
\label{tab:context}
\end{table}

Primary-fact acquisition requires inspection or comparison to return each
designated fact before the target's deadline. This differs from source
access ($O$), which can expose a relevant source without its required
fields. The input-visibility analysis then asks whether those acquired
facts remain available at the last valid decision.

Runs 2--3 retain 46,606 pre-action requests or bridge payloads across
96 trajectories: 41,285 API requests and 5,321 Astra bridge inputs. Every
retained action in this mapping has an associated input, including actions
from recovered continuation segments.
The API cohort contains all 80 trajectories from Runs 2--3 and 1,920 targets,
including 1,295 failures. Astra's 16 later local runs remain a separate
execution cohort.

For each target, we search the actual non-system input for each designated
record's primary fact, requiring the associated object or matter identifier.
Matching normalizes case and whitespace; a catalog listing alone does not
qualify. We use the last request within the target's valid window so that
later evidence cannot count as information available before its deadline.

\begin{table}[t]
\centering\small\setlength{\tabcolsep}{3pt}
\begin{tabular}{lrrr}
\toprule
System & Targets & $Y=0$ & Visible \\
\midrule
DeepSeek & 384 & 211 & 158 \\
Fable & 384 & 253 & 71 \\
Gemini & 384 & 280 & 62 \\
Grok & 384 & 301 & 107 \\
Qwen & 384 & 250 & 62 \\
\bottomrule
\end{tabular}
\caption{Input visibility across all eight worlds in Runs 2--3: 384 target instances per API system. Visible counts failures with all designated primary facts in the last input within the valid window.}
\label{tab:context-model}
\end{table}

At the last valid-window request, 460 API failures retain matches for all
designated primary facts and their matter identifiers.
Excluding all stage-uncertain targets leaves 1,545 targets, 920 failures,
and 310 last-window matches (33.7\%).

\section{Model Differences and Strategy Reuse}
\label{sec:paths}

\subsection{Full-Population Failure Classification}
\label{sec:failure-census}

\paragraph{Population and unit.}
\looseness=-1
The review covers all 2,096 target instances with $Y=0$ in the reported
scoring snapshot: 1,875 from the five API systems and 221 from Astra (Codex).
One instance is a model--run--world--target combination. We set aside
160 instances with normal constraints from the model-error classification,
leaving 1,936 instances: 1,524 with a specific observed problem and 412
with insufficient attribution evidence. Main-result denominators remain unchanged; classification does
not itself alter outcome scores. Table~\ref{tab:failure-census-coverage} lists exclusions
and evidence coverage for each system.

\paragraph{One primary observed problem per instance.}
\looseness=-1
LLM reviewers produced the initial classifications from the existing
reviews and adjudications. Human reviewers checked each classification
against the original actions and feedback, resolved disagreements, and
finalized the labels.
The primary label identifies the documented obstacle still blocking the
target. When several problems appear, an explicit unsupported prerequisite
can explain subsequent inaction, while a repaired parameter error does not
displace a remaining obstacle. Each instance receives one primary label;
the five problem types are:
\begin{itemize}[nosep]
\item \textbf{Self-restriction:} an unsupported prerequisite or expanded
prohibition is explicitly used to wait, refuse, or restrict progress.
\item \textbf{State or updating error:} a wrong matter, date, source, or
status is used in a decision, or a superseded judgment continues to govern it.
\item \textbf{Action selection or adjustment:} the record establishes a
permitted opportunity, but the agent fails to take the available action,
retry, follow up, or try another approach.
\item \textbf{Execution error:} an identifiable agent-side parameter,
binding, ordering, or procedure error obstructs the selected operation and
remains relevant to progress.
\item \textbf{Information gap:} needed facts are not obtained, assembled,
or retained in time, without evidence for a more specific primary problem.
\end{itemize}

Insufficient evidence is a separate status retained in every category-share
denominator, so the reported shares describe the full classification set.
It includes missing decision records and ambiguous action or scoring
interpretations that prevent selection of a primary problem. Fixed quota
exhaustion or a genuine permission limit alone does not establish model
error. Conversely, a rejected request does not justify exclusion when a
documented model mistake or unused permitted alternative explains the blockage.

\paragraph{Evidence and review.}
Each classification retains its target identity, reason, source records,
and event references where applicable. Public rules and agent-visible
information establish the relevant constraints; private target definitions
do not show that the agent had access to the same information.

User authorization does not by itself
resolve conflicting objective facts. A completed procedure can therefore
still contain an unsupported state choice. Outcome scoring follows the
simulated feedback under the target criteria in Section~\ref{sec:criteria}.

The initial synthesis combines model identities, offline agent reviews,
and recorded adjudications.
The full-confirmation review log records separately launched
\texttt{gpt-6.1-sol} contexts with \texttt{fork\_context=false} for its
assigned evidence checks. Retained review submissions and source-grounded
adjudications inform the final labels.

\begin{table*}[t]
\centering\small
\setlength{\tabcolsep}{4pt}
\begin{tabular}{lrrrrrrr}
\toprule
System & $N$ & \shortstack{Information\\gap} & \shortstack{Action\\selection /\\adjustment} & \shortstack{Self-\\restriction} & \shortstack{State /\\update\\error} & \shortstack{Execution\\error} & \shortstack{Insufficient\\evidence} \\
\midrule
DeepSeek & 296 & 55 & 48 & 61 & 41 & 5 & 86 \\
Fable & 367 & 170 & 37 & 86 & 18 & 1 & 55 \\
Gemini & 364 & 128 & 85 & 34 & 23 & 10 & 84 \\
Grok & 373 & 51 & 178 & 83 & 14 & 3 & 44 \\
Qwen & 325 & 81 & 42 & 97 & 24 & 20 & 61 \\
Astra (Codex) & 211 & 99 & 22 & 5 & 3 & 0 & 82 \\
\midrule
Total & 1936 & 584 & 412 & 366 & 123 & 39 & 412 \\
\bottomrule
\end{tabular}
\caption{Counts for the mutually exclusive primary-problem classification. Each row sums to $N$; insufficient evidence is retained as a separate status.}
\label{tab:failure-census-status}
\end{table*}

\begin{table*}[t]
\centering\small
\begin{tabular}{lrrrrr}
\toprule
System & All $Y=0$ & \shortstack{Normal\\constraints} & $N$ & \shortstack{Specific\\problem} & \shortstack{Insufficient\\evidence} \\
\midrule
DeepSeek & 298 & 2 & 296 & 210 & 86 \\
Fable & 368 & 1 & 367 & 312 & 55 \\
Gemini & 399 & 35 & 364 & 280 & 84 \\
Grok & 440 & 67 & 373 & 329 & 44 \\
Qwen & 370 & 45 & 325 & 264 & 61 \\
Astra (Codex) & 221 & 10 & 211 & 129 & 82 \\
\midrule
Total & 2096 & 160 & 1936 & 1524 & 412 \\
\bottomrule
\end{tabular}
\caption{Classification coverage. Normal constraints are excluded only from the error-classification denominator. Within that denominator, specific problems and insufficient evidence sum to $N$.}
\label{tab:failure-census-coverage}
\end{table*}

\paragraph{Scoring records.}
Both the main results and failure labels use the same finalized scoring records.
Unresolved stage-level scoring judgments retain their uncertainty flags. The
records include source hashes and action witnesses. The reproduction
procedure in Appendix~\ref{sec:reproducibility} uses these target-level records.

\paragraph{Astra: acknowledging uncertainty while choosing a date.}
In two returning-worker runs, Astra acknowledges an unresolved date
conflict but selects and executes one date. The active records state Day 21
and Day 26. In Run 2, the Day-9 confirmation request explicitly says
that neither consultation establishes which date is correct, while its
proposed action asks to use Day 21 and align the portal to it. On Day 12,
the agent executes that confirmation-bound action. Run 3 similarly states
that the consultations do not establish which date governs, then proposes
and executes the Day-21 choice (Run 2 indices 59 and 88; Run 3 indices
62 and 95 in the corresponding returning-worker timelines).
User approval authorizes a change, but does not establish which conflicting
date is correct: the agent chooses a date while acknowledging that the
evidence does not settle the conflict.

\subsection{Stage Profiles and Strategy Reuse}

\begin{table*}[t]
\centering\small
\begin{tabular}{lrrrrrrr}
\toprule
Target type & $n$ & $O$ & $H$ & $G$ & $T$ & $I$ & $Y$ \\
\midrule
Cross-day accumulation & 480 & 446 & 316 & 303 & 302 & 153 & 153 \\
Cross-source conflict & 495 & 465 & 308 & 277 & 268 & 172 & 172 \\
Missing expected event & 465 & 434 & 324 & 302 & 299 & 276 & 264 \\
Preference / multiple parties & 525 & 510 & 418 & 297 & 289 & 197 & 94 \\
Single-source constraint & 435 & 314 & 233 & 230 & 226 & 143 & 143 \\
Stakeholder clarification & 480 & 349 & 271 & 265 & 261 & 234 & 179 \\
\bottomrule
\end{tabular}
\caption{Stage counts by target type across five API systems and three runs. Each row pools the same target instances across all six criteria.}
\label{tab:stage-types}
\end{table*}

\paragraph{Models differ in where they fall short.}
The stage counts distinguish failure to recognize a problem from failure
to act on a timely judgment. Gemini has 180 cases with accessed evidence
but no specific hypothesis. Grok has 128/289 timely judgments without eligible delivery
(44.3\%), compared with 121/450 (26.9\%) for DeepSeek.

Multi-party targets expose a further gap between recognition and outcome.
DeepSeek's high overall observation and recognition coverage (556 and 497
of 576) accompanies its larger outcome count. On multi-party targets,
however, it recognizes 103/105 risks while grounding 79 and achieving 22
outcomes. Fable completes 45/96 clarification targets and 17/105
multi-party tasks.

\paragraph{Task pathways after timely discovery.}
\label{sec:task-paths}
To compare completion after recognition, we retain API instances with
$T=1$ and no uncertain stage judgments. All retained instances also have
timely primary facts. Success counts are 264/291 for missing expected
events, 143/209 for single-source constraints, 153/254 for cross-day
accumulation, 172/231 for cross-source conflicts, 179/232 for stakeholder
clarification, and 94/246 for preference or multi-party tasks. In particular,
obtaining and understanding the required evidence does not make single-source
tasks the easiest to complete.

\looseness=-1
The construction links these evidence patterns to different action routes:
missing-event tasks use a responsible-party resolution route, while
cross-day tasks require confirmed execution. The comparison therefore
describes complete task pathways rather than isolating a memory effect.
In Grok's Run-1 retired-tenant trajectory, a request to the responsible
institution completes a benefit-renewal matter on Days 10--11 (target
\texttt{01-02}, trace indices 80, 84). For a tax-declaration matter
(\texttt{01-03}), it compares the records and recognizes the conflict on
Day 14 (107--112), but delivers no qualifying intervention by Day 23.
These two instances have timely primary facts and grounded judgments,
with no uncertain stage labels, yet differ in outcomes after discovery.

\paragraph{Coverage differences matter for model comparisons.}
Direct inspection has a horizon quota of 46, whereas each world releases
68--71 records. Comparison can return new facts without another direct
inspection. Table~\ref{tab:shared-fact-outcomes} compares outcomes on the
same 311 target instances whose primary facts all five API systems obtain
in time. Removing targets with any system's review uncertainty leaves
139 instances. Relative outcome counts change with this shared-access
selection. This comparison measures success with shared fact access;
the full-cohort ranking also captures fact acquisition.

\begin{table}[t]
\centering\small
\begin{tabular*}{\columnwidth}{@{\extracolsep{\fill}}lrr@{}}
\toprule
System & All (311) & Unambiguous (139) \\
\midrule
DeepSeek & 161 & 110 \\
Fable & 179 & 103 \\
Gemini & 142 & 87 \\
Grok & 121 & 70 \\
Qwen & 184 & 98 \\
\bottomrule
\end{tabular*}
\caption{Target successes on shared timely-fact subsets. Both columns
require all five systems to obtain the primary facts; the second also
excludes any system's uncertain criterion judgments.}
\label{tab:shared-fact-outcomes}
\end{table}

The 239-instance DeepSeek--Grok comparison requires timely fact acquisition
by both systems and excludes targets with either system's uncertain
criterion labels. Qualifying interventions number 185 and 127, respectively.
Thus, the delivery gap is also present among instances where both systems
obtain the designated facts and neither has an uncertain criterion label.

\paragraph{What stage separation reveals.}
Stage combinations describe breaks in progress and may overlap; they are
distinct from the mutually exclusive primary-problem labels.
Across the five API systems, 652 of 2,880 target instances have
$O=1,H=0$, showing that source access can stop short of recognizing a
specific risk. Another 470 have $T=1,I=0$: a timely, grounded judgment
does not lead to a qualifying intervention. The gap after delivery is
concentrated in preference, authorization, and multi-party targets:
these account for 103 of the 170 cases with $I=1,Y=0$ (60.6\%) but only
525 of 2,880 instances (18.2\%). Within these task types, 418 risk judgments
(79.6\%) yield 94 outcomes (17.9\%).

Table~\ref{tab:stage-types} separates evidence and action dependencies by
target type. For single-source, cross-source, and cross-day targets, all
qualifying delivered interventions have matching outcomes in the observed
runs: their combined 1,410 instances contain no $I=1,Y=0$ case. Missing-event,
clarification, and preference/multi-party targets contribute 12, 55, and 103
such cases, respectively. The criteria expose where task pathways differ.

\paragraph{Checkpoint evidence for strategy reuse.}
\label{sec:strategy-reuse}
The fixed stratified sample contains two randomly selected failures per
API system and world, for 80 targets across five systems and eight worlds.
The sample was drawn from the 1,739 failed target instances in the
111-trajectory screening cohort. The review combines
target traces, archived judgments, input checks, and checkpoint replays.
Each replay uses the corresponding public object identifiers and remaining
comparison allowance; the original day-close operation is also reproduced.
Each replay tests whether the missing facts could be acquired from that
target's original state using the resources still available.

Among the 35 targets missing primary facts,
15 have successful comparison replays at checkpoints 5--10 days before
their deadlines: six Fable, three Gemini, two Grok, and four Qwen cases.
Five of these trajectories previously acquired fresh facts by comparison
(three Fable and two Gemini cases).

\begin{table}[t]
\centering\small
\begin{tabular}{lrrr}
\toprule
System & Cases & Days left & Comparisons left \\
\midrule
Fable & 6 & 6--10 & 11--16 \\
Gemini & 3 & 7--10 & 9--20 \\
Grok & 2 & 9--9 & 17--20 \\
Qwen & 4 & 5--9 & 11--22 \\
\bottomrule
\end{tabular}
\caption{Original-checkpoint fact-acquisition replays: successful cases, days until deadline, and remaining horizon comparisons. Every checkpoint has exhausted new-object inspections but permits a daily comparison.}
\label{tab:replay-feasibility}
\end{table}

Table~\ref{tab:replay-feasibility} records the resources at these checkpoints.
All 15 have exhausted the 46-object inspection allowance but retain an
unused daily comparison slot and 9--22 remaining horizon comparisons. Direct access
fails under the original quotas; comparison returns the missing primary
facts 5--10 days before the deadline. It is an information-only operation
with no user commitment. For example, Fable's contractor run has 14
comparisons left on Day~49, before a Day~57 deadline. Comparing the two
public object IDs returns the conflict between a requested public showreel
and an NDA forbidding public excerpts. The same trajectory had obtained
fresh facts through comparison on Day~40.

The replay endpoint is fact retrieval at each target's original checkpoint.
The 15 successful replays therefore establish individually available
alternatives, with resources evaluated separately at each checkpoint.
Aggregate counts, checkpoint resource
summaries, and source hashes are in \path{results/acquisition_adaptation.json}
and \path{results/replay_feasibility.json}.

\subsection{Astra and DeepSeek: Discovery to Action}
\label{sec:astra-follow-through}

\paragraph{Matched timely judgments.}
We match instances by run, world, and target, retaining those with $T=1$
and no uncertain stage judgments for either system. This yields 275
instances: Astra has 268 qualifying interventions and 262 successes,
versus DeepSeek's 227 and 207. Both systems have already made timely,
evidence-supported judgments on these same instances, placing the
remaining difference in intervention and completion. These counts complement
Table~\ref{tab:mean-stages}'s full-cohort rates.

\paragraph{A shared discrepancy, different next steps.}
In Run 2 of the returning-to-work caregiver world, both systems inspect a claimant
commitment allowing 24 hours of availability on Day 3 and a portal
assignment of 35 search hours on Day 7. Both recognize the conflict.
Astra requests confirmation on Day 7 to submit a correction request,
while preserving both records and leaving factual reconciliation to the
responsible record holder. It executes the confirmed action on Day 8;
Day-9 environment feedback confirms option preservation, not resolution
of the conflicting records. DeepSeek saves the discrepancy and registers
a watchpoint on Day 7, waiting for the record holder to become available,
but delivers no qualifying intervention for this target.
The difference is how recognized uncertainty guides the next step:
Astra uses an authorized protective action without claiming to have
settled the facts; DeepSeek retains the conflict as a monitoring condition.
The archived target is \path{risk-caregiver_returning_to_work-01-02}.
Evidence appears at Astra trace indices 17, 41, 44, 49, and 56, and
DeepSeek indices 24, 61, 71, and 72; indices are zero-based in the Run-2
traces mapped by \path{main_manifest.json} in the reproduction archive.

\paragraph{What a harness explanation would test.}
Both setups have a \emph{harness}: the execution framework that supplies
context, dispatches tools, and returns feedback to the model.
The common runner starts a new conversation each day, carrying saved
memory and watchpoints; Astra uses the local Codex agent interface.
Codex's documented harness manages conversation state and tool execution
across turns \citep{bonamy2026codexharness}.
This makes continuity of pending steps and use of action feedback plausible
contributors to follow-through. The observed comparison changes both the
model and its harness, so it describes their combined behavior. A
same-model comparison across harnesses would isolate whether these
execution features improve follow-through; the case above identifies
concrete behaviors to test: turning a recognized conflict into an
authorized action and checking its outcome.

\subsection{Paired Trajectories After Recognition}
\label{sec:paired-actions}

The following comparisons link current outcome labels to actions in the
archived traces. Each cited instance has timely primary facts, a timely
grounded judgment, and no uncertain stage labels. Trace indices are
zero-based and include feedback events; \path{main_manifest.json} maps
each run, system, and world to its archived trace.

\paragraph{Grok and DeepSeek: monitoring versus authorized execution.}
In Run 3 of the food-business world, a supplier credit memo conflicts
with a scheduled full debit (target \path{risk-neighborhood_food_business_owner-03-02},
deadline Day 32). Grok obtains the records on Days 12 and 24, looks up the
confirmation-based procedure, then saves the discrepancy, a watchpoint,
and a risk report on Day 24 (indices 81, 170--176). DeepSeek obtains the
same facts (141, 328), requests confirmation on Day 24 (332), receives it
on Day 25 (339), and executes on Day 27 (384), followed by preservation
feedback on Day 28 (397). Across the three runs, Grok never completes this
target while DeepSeek completes it each time. Grok's surface behavior
varies: Run 1 acts after the deadline, Run 2 records a reminder without
an external state change, and Run 3 monitors the conflict. Different
action sequences can thus yield the same persistent failure.

\paragraph{Fable and Qwen: equal totals, opposite task outcomes.}
Across 576 matched instances, both succeed in 132, only Fable in 76,
only Qwen in 74, and neither in 294. Requiring timely primary facts and
no uncertain stage labels for both leaves 249 instances: 132 joint
successes, 37 Fable-only, 41 Qwen-only, and 39 joint failures. Thus, both
systems obtain timely facts in 78 of the 150 full-cohort disagreements.

Two Run-1 caregiver tasks illustrate the two directions. The target IDs
in this world share the prefix \path{risk-caregiver_returning_to_work};
the suffixes below identify each matter. For the care allowance
discrepancy (\texttt{03-02}),
Fable consults both parties on Days 25--26, requests confirmation on
Day 27, and follows up on Day 29 after no response. Confirmation and
execution occur on Day 30, before the Day-37 deadline (indices 228--285).
Qwen's confirmation request precedes completed consultation and is
rejected; later contact attempts are also rejected. It saves a condition
requiring fresh instructions before retrying and does not complete the
target (319--323, 374--376, 407, 508).

The direction reverses for a childcare funding code that covers autumn
but appears against a summer-session charge (\texttt{01-03}). Fable recognizes the
mismatch on Day 13 and waits for another record or invoice amount before
contacting the provider (97--102). Qwen contacts the provider on Day 13,
follows up after a partial reply on Day 16, receives clarification on
Day 17, and intervenes on Day 18 before the Day-22 deadline (166--257).
Both trajectories finish 12 of 24 targets; their identical totals conceal
different choices about follow-up on these two matters.

\paragraph{DeepSeek across runs: stable discovery, variable next steps.}
For caregiver target \texttt{03-03}, all three runs identify
a pension statement recording seven years of service and a conflicting
zero-service quote, with a Day-36 deadline. Runs 1 and 2 use the reversible
procedure to preserve the service record without deciding which entry
governs: interventions occur on Days 27 and 28, with next-day feedback
(Run 1 indices 348, 351, 352, 360; Run 2 indices 363, 365, 378, 386).
Run 3 obtains the discrepancy and the same available procedure on Day 27
(388, 390), but saves a watchpoint awaiting another record (394).
Its Day-36 memory still treats the matter as pending (506), and no
qualifying intervention occurs. Here repeat inconsistency lies in using
an available protective action after consistently recognizing the problem.

\paragraph{DeepSeek's mortgage case: accepted execution, different outcomes.}
The 122-instance coordination subset requires timely primary facts and
grounded judgment, no uncertain stage labels, completed consultations with
both required parties, a user-approved action, and accepted execution by
the deadline. Of these instances, 94 succeed; among the 28 failures, 21
receive intervention credit but not outcome credit, and seven receive
neither. Accepted execution and a qualifying intervention are therefore
separate checks.

The mortgage example is target \texttt{03-02} in the
\path{midcareer_worker_after_layoff} world. In Run 1, DeepSeek proposes
asking the loss-mitigation office to apply
the hardship pause on Day 26 (356), obtains confirmation and executes on
Day 27 (366, 370), with feedback on Day 28 (386). In Run 2 it instead
proposes treating the letter as controlling and pausing the debit itself
on Day 27 (395), then obtains confirmation and executes on Day 28
(408, 410). Although execution is accepted and feedback arrives (434),
the action does not satisfy the target's requirements. Both runs complete
consultation and authorization; the outcome differs with the proposed
resolution of the conflicting payment instructions.

\paragraph{Explicit feedback supports local revision.}
\label{sec:feedback-repair}
We identify each API target instance's first material-return or
partial-resolution feedback, using the operation or matter identifier and
linked transaction IDs to associate events with the target. Eligibility
requires that the feedback arrive by the deadline and that its specified
follow-up day be no later than that deadline. It does not require a later
request or successful outcome. An accepted follow-up is a subsequent
contact on or after the allowed day and by the deadline, explicitly citing
the feedback ID and receiving a \texttt{request\_logged} response.

Of 102 material-return and 134 partial-resolution instances, 99 and 126
meet these timing conditions. All eligible instances have an accepted
follow-up; the current target outcomes are successful in 91/99 and
109/126, respectively. Restricting further to timely primary facts,
$T=1$, and no uncertain stage labels yields 95 and 113 instances, again
all with accepted follow-ups, and 91 and 109 successes. These feedback
messages specify the missing \texttt{case\_reference} field, the revision
format, and the requirement to bind its value to an inspected source.
The result characterizes repair within these explicitly guided interactions.

In Grok's Run-1 contractor trajectory, the licensing correction target
\texttt{01-02} illustrates the sequence. On Day 8, Grok receives notice
that user confirmation is unnecessary and switches to contacting the
responsible desk (indices 68--69). The Day-9 return specifies the missing
field (73). Grok rereads the records on Day 10 (84--85), submits the
field value with its source and feedback reference (92), and receives
resolution feedback on Day 11 (100). Its current outcome label is
successful and has no uncertain stage judgment. This successful repair
distinguishes responding to explicit correction instructions from deciding
independently when and how to restart unfinished work.

\section{Resource Use and Temporal Patterns}
\label{sec:resources}
\label{sec:resource-observation}

\paragraph{More calls can maintain pending work rather than complete it.}
Across all three runs and eight worlds, Fable makes 8,049 tool calls for
208 successes and Qwen makes 12,553 for 206. Table~\ref{tab:tool-allocation}
breaks down every counted call by tool. Counts include attempted calls,
errors, and day-close calls; they exclude environment feedback, protocol
recovery events, and forced day closures. These are tool attempts,
not provider requests or costs.

\begin{table}[t]
\centering\small
\begin{tabular*}{\columnwidth}{@{\extracolsep{\fill}}lrr@{}}
\toprule
Tool & Fable & Qwen \\
\midrule
Read source & 1,252 & 1,687 \\
Inspect object & 1,166 & 1,920 \\
Compare objects & 259 & 241 \\
Look up procedure & 351 & 1,195 \\
Report risk & 286 & 1,337 \\
Register watchpoint & 713 & 1,629 \\
Save memory & 1,603 & 1,741 \\
Contact stakeholder & 370 & 621 \\
Request confirmation & 249 & 330 \\
Execute confirmed intervention & 108 & 139 \\
Execute reversible intervention & 253 & 354 \\
Finish day & 1,439 & 1,359 \\
\midrule
Total & 8,049 & 12,553 \\
\bottomrule
\end{tabular*}
\caption{Tool-call allocation across 24 trajectories per system. Attempts
are counted whether accepted or rejected; counts are distinct from
qualifying interventions and successful target outcomes.}
\label{tab:tool-allocation}
\end{table}

Risk reporting and watchpoint registration account for 1,967 of Qwen's
4,504 additional calls (43.7\%). Procedure lookups add 844 and inspections
754; the remaining tools together add 939. Reporting and monitoring help
retain unresolved matters, but their repetition does not itself deliver
the next external action. Qwen also makes more contact and intervention
attempts, showing why call counts must be read alongside acceptance,
timing, and target outcomes. Fable uses fewer calls in all 24 matched
world--run pairs, yet has more successes in nine, fewer in ten, and the
same number in five. The resource difference is consistent across these
pairs, whereas outcome superiority is not.

\paragraph{Resource totals and later tasks.}
In Run 3, Qwen uses 196.6 million recorded input tokens and 4,257 tool
calls for 76 successes; DeepSeek uses 145.6 million and 4,087 calls for 109.
This cross-model comparison aggregates all eight worlds; input totals
include retransmitted context. These totals describe resource use and
outcomes under each system's model and action policy. Higher resource use
can accompany lower success; the number of
interactions alone does not indicate whether they advance unfinished tasks.

We next examine when tasks become detectable and whether failures follow
budget exhaustion. We group targets by earliest detection day: early (Days 1--20), middle
(21--40), and late (41--60). Late-target success is lower than early-target
success in 14 of 15 model--run groups. DeepSeek Run 3 is the
exception, with 31/59 early-target successes and 28/46 late-target successes.

\paragraph{Inspection capacity is depleted before late targets appear.}
Counting successful budget-consuming inspections, rather than rejection
messages, shows that all 120 API trajectories use their 46 direct
inspections by Days 31--39. Free retrievals of previously inspected records
do not enter this count. All 690 late-onset API instances therefore become
detectable after this allowance is exhausted. Yet 88 of the 120 trajectories
record no horizon-level read/inspect rejection: an agent that stops trying
after exhaustion produces no such error. For example, Qwen's Run-1
laid-off-worker trajectory consumes its 46th inspection on Day 32 (index
463) and records on Day 42 that newly arrived records cannot be inspected
with the remaining allowance (575), without a horizon-level rejection.

Gemini records such rejections in 16/24 trajectories; targets appearing
after its first rejection have 8/112 outcomes, versus 169/464 otherwise.
This is an association with attempts made after depletion, not a division
between exhausted and unexhausted trajectories. Direct inspection and
other operations have separate budgets: record comparison provides another
route to new facts, as the checkpoint replays in
Appendix~\ref{sec:strategy-reuse} demonstrate.

\paragraph{Late failures differ in fact-acquisition coverage.}
Each API system has 177 early and 138 late target instances across its
three runs. All 177 early instances have timely primary facts in each
system. Among late instances, DeepSeek obtains those facts for 108/138,
compared with 12/138 for Fable, 19/138 for Gemini, 6/138 for Grok, and
3/138 for Qwen. The latter four systems lack timely primary facts in
512 of 526 late failures (97.3\%); excluding any uncertain stage labels
leaves 372 of 381 (97.6\%). DeepSeek lacks them in 30 of its 77 late
failures. Thus much of the late deficit in those four systems is located
at fact acquisition, whereas DeepSeek more often reaches a later stage.
The time comparison spans changing information-access conditions and task
mixes; it does not isolate deterioration in memory. Fact availability still
does not ensure completion: in DeepSeek's Run-1 laid-off-worker trajectory,
target \texttt{06-01} has its primary facts by Day 42 (index 576), before
the Day-49 deadline, but receives $T=1,I=0,Y=0$ under the current criteria.

\section{Information and Action Follow-ups: Protocol and Checks}
\label{sec:recovery-protocol}

These experiments test recovery at selected natural-failure checkpoints:
task completion after information support, and timely intervention and
task completion after action reminders.

\begin{table*}[t]
\centering\small
\begin{tabular}{@{}lccc@{}}
\toprule
\multicolumn{4}{@{}l}{\textbf{(a) Information support: original target outcome, success / failure / unknown}} \\
Checkpoint (Gemini) & Original & Added facts & Reformatted facts \\
\midrule
Buyout (R2) & 0 / 3 / 0 & 3 / 0 / 0 & 2 / 1 / 0 \\
Training (R2) & 0 / 2 / 1 & 0 / 1 / 2 & 0 / 0 / 3 \\
Training (R3) & 0 / 3 / 0 & 3 / 0 / 0 & 2 / 0 / 1 \\
\bottomrule
\end{tabular}
\par\medskip
\begin{tabular}{@{}lcccc@{}}
\toprule
\multicolumn{5}{@{}l}{\textbf{(b) Action reminders: timely new action and original target outcome}} \\
& \multicolumn{2}{c}{General review} & \multicolumn{2}{c}{Action review} \\
Case (original model) & New action & Outcome & New action & Outcome \\
\midrule
Scheduling (Gemini) & 0/3 & 0/3 & 1/3 & 0/3 \\
Car (DeepSeek) & 0/3 & 0/3 & 1/3 & 1/3 \\
\bottomrule
\end{tabular}
\caption{\textbf{Information support can restore task success; action reminders can restore action without timely completion.} Three continuations per checkpoint and condition, all through Day 60. (a) Each cell totals three; unknown scores remain in the denominator. R2/R3 identify source runs, not new repeats. (b) New action uses the original timely, grounded intervention criterion ($I$); outcome uses the original target criterion ($Y$). The two training checkpoints share one world--target pair. In (a), new-$I$ and $Y$ counts coincide; the criteria remain distinct.}
\label{tab:recovery-followups}
\end{table*}

\paragraph{Information outcomes across conditions.}
All 18 supported continuations receive the designated primary facts:
ten succeed, two are confirmed failures without a qualifying next step,
and six have uncertain outcome scores. At Run-2 training, the six supported
continuations yield one failure and five unknowns. Reformatting yields
2/3 successes at both responsive checkpoints, versus 3/3 with added facts.
Seven outcomes remain unknown
across all 27 information continuations.

\subsection{Checkpoints and Selection}

Both experiments resume archived natural-failure states, generating fresh
continuations for every condition, including the controls.
Table~\ref{tab:recovery-roster} identifies
the five checkpoints. All 39 scheduled continuations reach Day 60;
none stops upon focal success. Target windows, permissions, budgets,
model routes, and scoring rules are unchanged.

\begin{table*}[t]
\centering\small
\begin{tabular}{@{}llllrr@{}}
\toprule
Experiment & Checkpoint & World / focal matter & Model (run) & Start day & Deadline \\
\midrule
Information & \texttt{4c9bfe779e6bb394} & Creative contractor / buyout & Gemini (2) & 46 & 55 \\
& \texttt{b71e084171730d98} & Laid-off worker / training hours & Gemini (2) & 48 & 57 \\
& \texttt{7f10db11d4df4f8e} & Laid-off worker / training hours & Gemini (3) & 48 & 57 \\
Action & \texttt{ca3a58a5edf2b82c} & Food business / inspection and pickup & Gemini (2) & 15 & 22 \\
& \texttt{05cdc677c2bc1e7d} & Multigenerational household / shared car & DeepSeek (2) & 16 & 23 \\
\bottomrule
\end{tabular}
\caption{Frozen natural-failure checkpoints. The deadline is the latest
qualifying action day, not the end of the continuation; later feedback is
checked against the action it confirms. The two training checkpoints
test the same target in the same world.}
\label{tab:recovery-roster}
\end{table*}

\paragraph{Information selection.}
The roster consists of three previously qualified retrieval paths, fixed
before new outputs. Each checkpoint lacks primary facts that a permitted
native comparison can retrieve. The two training checkpoints share one
world--target pair; together with the buyout case they form two clusters.
This purposive selection tests targeted acquisition of previously
unobserved facts and whether supplying them restores progress on the
original task.

\paragraph{Action selection.}
A frozen screen starts from 412 records carrying the action-selection
label in the screening snapshot. Recorded descriptions assign provisional
strata for omission, follow-up/retry, and route adjustment; retrieval-related
and unclassified descriptions do not enter those strata. A stratified hash
rule selects 12, with at most one per source trajectory, before checking
the archived state for a remaining permitted opportunity.

Admission requires six conditions at the exact restart boundary: the
relevant facts are present in the decision input; the agent has expressed
the correct interpretation; its understanding of the current state and
permissions is correct; a substantive permitted next action is available;
no unsupported restriction is driving the pause; and sufficient time and
resources remain. These checkpoint conditions isolate the opportunity
for an action reminder. The historical classification concerns the
trajectory's observed problem, whereas admission concerns the state from
which the experiment will resume.

Two candidates meet all six conditions. For nine, the archived input and
pre-action state cannot be jointly bound, or the available record does not
establish the required interpretation, permission, or remaining action
window. One is excluded because the agent explicitly infers a
private-reminder-only restriction from the process response, placing an
unsupported permission judgment upstream of inaction.
The two-case roster is fixed before continuation outputs are generated.
The food-business case concerns following up when
a contact is available; the car case concerns progress after daily
capacity returns. Further process lookups may still be needed to identify
the procedural fields required for the next action.

\subsection{Interventions and Resource Accounting}

\paragraph{Information support.}
The \emph{added facts} and \emph{reformatted facts} conditions
correspond to raw support and transposed support, respectively,
in the archived protocol. We use the descriptive names throughout the tables.
Original continuations receive no added evidence. Both support conditions
execute one researcher-selected, read-only \texttt{compare\_objects}
operation on the relevant public record pair. The added-facts condition
supplies the returned facts and a literal table. The reformatted-facts
condition exchanges the table's rows and columns without changing its cells.
Neither condition explains the discrepancy or supplies a plan for the next action.

Normal comparison and daily-action charges apply; the two support conditions
share the resulting state.
Additional prompt tokens are charged; researcher selection effort is not
included in model cost. The primary added-facts versus original contrast tests the
package of targeted retrieval, evidence availability, salience, and its
resource use. The secondary reformatted-facts versus added-facts contrast
tests only presentation.

The acquired evidence is presented as environment-provided information
in a user message. Appending this evidence leaves earlier messages and
model signatures unchanged.

\paragraph{Action reminders.}
Each reminder is appended once as a user message at the admitted boundary;
the native next-day reset remains unchanged. Both share this opening instruction:
\begin{quote}\small
Review unfinished matters using the information currently available,
remaining resources, and existing authorization limits. This reminder
adds no facts, permissions, or resources.
\end{quote}
General review continues:
\begin{quote}\small
Consider the current situation and unresolved matters, assess what remains
appropriate under existing conditions, then choose your next step.
\end{quote}
Action review instead continues:
\begin{quote}\small
Check whether permitted follow-up, retry after a lifted restriction, or
another lawful approach is appropriate, then choose your next step.
\end{quote}
The comparison estimates the effect of action review relative to general
review at each checkpoint; general review is the control condition.
Results remain separate for each model--case combination.

\subsection{Scoring and Uncertainty}

\paragraph{Original endpoints.}
The information experiment's primary endpoint is the original target
outcome $Y$; the action experiment's is a new post-checkpoint timely,
grounded intervention $I$, with $Y$ secondary. The controlled information
comparison never counts as an autonomous qualifying intervention.
Success requires the original evidence, timing, authority, and outcome
checks, not merely an accepted tool call. $Y$ can mean preserving an option
pending resolution, even while the underlying issue awaits a final resolution.

Each of the 39 joined prefix--continuation histories is evaluated by two
separately launched AI review contexts, both configured as
\texttt{gpt-6.1-sol} with high reasoning effort and
\texttt{fork\_context=false}. Acting-model and condition labels are withheld;
neither context receives the other's judgments. Source checks by the
organizing AI agent and the original scoring program follow. The
intervention text itself remains visible in the history. These 78 accepted
AI reviews each check all 24 targets and post-checkpoint authority attempts.
The final review pairs are sealed, with source hashes checked for each pair.

\paragraph{Seven unknown information outcomes.}
Six involve whether a training-hours request merely asks an authority to
reconcile 400 versus 720 hours, or assumes eligibility for 720 without
establishing the required supervisory proof. The remaining case concerns
whether a multi-target daily summary establishes focal risk recognition
($H$). The frozen dependency rules preserve this upstream uncertainty.
All seven remain in the fixed denominators. The action experiment has
resolved judgments for every focal $I$ and $Y$ outcome.

\paragraph{Analysis units.}
Per-checkpoint counts are the primary presentation. The prespecified
information summary first equally weights checkpoints within a
world--target pair, then equally weights the two pairs. Assigning unresolved
endpoints their worst and best values on the fixed continuations bounds
each contrast. Added facts
exceeds original continuation by $+66.7$ to $+91.7$ percentage points;
the increment from reformatting ranges from $-41.7$ to $+8.3$ points.
Three repetitions characterize within-checkpoint variation. Information and action results
describe recovery at the selected checkpoints.

\subsection{Execution and Secondary Checks}

\begin{table*}[t]
\centering\footnotesize
\begin{tabular*}{\textwidth}{@{\extracolsep{\fill}}llrrrr@{}}
\toprule
Checkpoint & Condition & Authority attempts & Actions & Other outcomes & Requests \\
\midrule
\multicolumn{6}{@{}l}{\textit{Information experiment}} \\
Buyout (R2) & Original & 0--13 & 319 & 24--35 & 371 \\
Buyout (R2) & Added facts & 0--0 & 75 & 24--33 & 118 \\
Buyout (R2) & Reformatted facts & 0--1 & 119 & 22--32 & 166 \\
Training (R2) & Original & 0--1 & 41 & 25--36 & 84 \\
Training (R2) & Added facts & 0--4 & 73 & 23--33 & 116 \\
Training (R2) & Reformatted facts & 0--2 & 23 & 24--36 & 67 \\
Training (R3) & Original & 0--2 & 10 & 26--42 & 49 \\
Training (R3) & Added facts & 0--0 & 17 & 26--40 & 58 \\
Training (R3) & Reformatted facts & 0--1 & 15 & 27--36 & 55 \\
\midrule
\multicolumn{6}{@{}l}{\textit{Action experiment}} \\
Scheduling (Gemini) & General review & 18--29 & 739 & 28--32 & 932 \\
Scheduling (Gemini) & Action review & 15--29 & 753 & 28--32 & 936 \\
Car (DeepSeek) & General review & 1--19 & 805 & 36--51 & 1230 \\
Car (DeepSeek) & Action review & 0--6 & 655 & 35--47 & 1078 \\
\bottomrule
\end{tabular*}
\caption{Secondary checks summed over three continuations per row.}
\label{tab:recovery-secondary}
\par\smallskip\begin{minipage}{\textwidth}\footnotesize
\textit{Authority attempts:} bounds run from confirmed unauthorized attempts to confirmed plus unresolved attempts; unresolved attempts are not confirmed violations.\par
\textit{Actions:} reviewed post-checkpoint actor actions, excluding controlled information acquisition.\par
\textit{Other outcomes:} bounds for 23 non-focal targets per continuation (69 instances per row); scores can vary with full-history labels.\par
\textit{Requests:} model HTTP requests, not successful tool calls.
\end{minipage}
\end{table*}

\paragraph{Complete histories, including unsuccessful actions.}
The information experiment contains 27 continuations and 1,084 requests;
the action experiment contains 12 continuations and 4,176 requests.
Result and cost totals cover these 39 analyzed continuations.
All 27 information-experiment
terminal states are accounted for, and recorded state replay passes.
Seven explicit no-action responses (three malformed-function responses and
four token-limit responses) remain in the record. Continuations use the
bounded protocol recovery and a 4,096-token output limit, retaining saved
responses.

\paragraph{What recovery looks like in the traces.}
\begin{itemize}[leftmargin=*,itemsep=3pt,parsep=0pt]
\item \textbf{Scheduling.}
In the scheduling action condition, a Day-22 inquiry about the
inspection--pickup overlap qualifies as new $I$. The agent executes the
user-approved resolution on Day 27 and receives feedback on Day 28.
Outcome credit depends on the protective operation's actual date,
which falls after the Day-22 deadline.
\item \textbf{Shared car.} On Day 19, the agent applies a reversible flag
to the overlap between the clinic pickup and the grandson's car-use period,
leaving both records unchanged. Day-20 feedback confirms that the pickup
option is preserved pending household coordination.
\item \textbf{Buyout.} In four continuations, a Day-46 process lookup
returns the authorized recipient's identifier. The recipient becomes
available on Day 47: contacting that identifier on Day 46 returns
\texttt{unknown stakeholder}, and retrying on Day 47 produces an accepted
request. Thus, discovering whom to contact precedes being able to contact
them. The accepted retry then provides the receipt required for delivery credit.
\end{itemize}

\paragraph{Authority and other targets.}
Table~\ref{tab:recovery-secondary} reports all conditions.
Information continuations have no confirmed
unauthorized attempts, but 24 remain unresolved across 692 reviewed actor
actions. Action continuations include confirmed unauthorized attempts in
both reminder conditions. Normal quota errors alone do not count as
authority violations.

For the other 23 targets, independent full-history reviews also reassess
unresolved judgments in the shared pre-intervention history; resulting
label differences need not reflect intervention effects. All prespecified
other-outcome contrasts have uncertainty ranges spanning negative and
positive values.

\paragraph{Costs and availability.}
Provider-reported information-continuation cost is US\$31.0364. The action
experiment reports US\$35.5638 for Gemini; DeepSeek has no provider-reported
charge and a frozen-tariff upper bound of US\$22.3435. These figures cover
actor-model requests in the analyzed continuations; researcher curation
and evaluation reviews are outside this cost scope.

Appendix~\ref{sec:reproducibility} describes the aggregate exports and the
full-record recovery archive, including access and public-copy transformations.

\section{Constraint Intervention: Protocol and Checkpoint Results}
\label{sec:permission-protocol}

\subsection{Design and Outcome Checks}

\textbf{Does correcting a supplied permission judgment restore an allowed inquiry?}
At six checkpoints from six worlds, Qwen3-Max receives one of three
externally supplied notes that
\emph{retain} the incorrect no-contact judgment, \emph{rewrite} it with the
same meaning, or \emph{correct} it to distinguish inquiry from commitment.
Rewriting tests whether new wording suffices; correction changes the
permission judgment. Within each presentation and permission setting, we
hold the world, evidence, tools, resources, and actual permissions fixed.

We also test whether subsequent updates restore action while the false
note remains. The \emph{current-state} and \emph{update} presentations
jointly vary record order and a temporary waiting instruction that is
explicitly released before action. Their exact input sequences appear below.

Each window is an independent interaction from a checkpoint. Six checkpoints,
three note forms, two presentations, two permission settings, and ten
repeats yield 720 windows. Half permit inquiry but prohibit commitments;
half also prohibit contact. We count accepted target-related clarification
inquiries to authorized recipients. Each plotted count pools six checkpoints
and ten repeats for a total of 60 independent windows per plotted count.

Section~\ref{sec:permission-experiment} and Figure~\ref{fig:followups}c
report the pooled results; Table~\ref{tab:permission-worlds} separates them
by checkpoint. Across the experiment, no preference-bearing commitment is
executed; two confirmation-required state-change attempts are rejected.

\subsection{Checkpoint Selection and Protocol}

\begin{table*}[t]
\centering
\small
\setlength{\tabcolsep}{4pt}
\begin{tabularx}{\textwidth}{@{}>{\raggedright\arraybackslash}p{0.25\textwidth}>{\raggedright\arraybackslash}Xrrrrrr@{}}
\toprule
 & & \multicolumn{3}{c}{Current state} & \multicolumn{3}{c}{Update} \\
\cmidrule(lr){3-5}\cmidrule(lr){6-8}
Original world & Matter & Retain & Rewrite & Correct & Retain & Rewrite & Correct \\
\midrule
Returning caregiver & Childcare funding & 3/10 & 7/10 & 10/10 & 10/10 & 10/10 & 10/10 \\
Newcomer coordinator & Credential account contact & 0/10 & 0/10 & 10/10 & 6/10 & 5/10 & 10/10 \\
Multigenerational household & Co-owner signatures & 0/10 & 0/10 & 10/10 & 6/10 & 6/10 & 10/10 \\
Creative contractor & Insurance timing & 1/10 & 0/10 & 10/10 & 9/10 & 10/10 & 10/10 \\
Retired tenant & Account contact authority & 1/10 & 1/10 & 10/10 & 10/10 & 10/10 & 10/10 \\
Returning worker & Payment-date/year-end rate & 0/10 & 0/10 & 10/10 & 10/10 & 8/10 & 10/10 \\
\midrule
Total & & 5/60 & 8/60 & 60/60 & 51/60 & 49/60 & 60/60 \\
\bottomrule
\end{tabularx}
\caption{Accepted clarification inquiries when contact is permitted but commitments are not. Each row is one checkpoint from a different original world, with ten repeats per cell. The total row supplies the six bars in Figure~\ref{fig:followups}c.}
\label{tab:permission-worlds}
\end{table*}

Table~\ref{tab:permission-worlds} separates the pooled counts in
Figure~\ref{fig:followups}c by checkpoint and condition.

Six checkpoints were selected before collecting new model outputs.
Eligibility required public evidence for a permitted inquiry and an
offline execution check of both its acceptance and the no-contact control.
Table~\ref{tab:permission-worlds} maps each checkpoint to its tested matter.
In four cases, this differs from the matter initially examined in the
source snapshot.

\paragraph{Historical notes.}
The common input presents the note as an earlier diagnostic judgment.
Notes have Boolean fields
\texttt{contact\_now} and \texttt{commitment\_now}, plus a summary.
The commitment field is false in every condition. Contact is false in the
retain and rewrite notes, and true in the corrected note when inquiry is
allowed. Each note also includes the following summary text:
\begin{itemize}[leftmargin=*,itemsep=3pt,parsep=0pt]
\raggedright
\item \textbf{Retain:} ``No contact permitted: approval for a
preference-bearing commitment is still missing. No commitment allowed
under the \mbox{current user directive}.''
\item \textbf{Rewrite:} ``Contact is not permitted because approval for a
preference-bearing commitment remains absent. Commitment is not allowed
under the current user directive.''
\item \textbf{Correct:} ``Clarification contact is permitted: it is not a
preference-bearing commitment. No commitment allowed under the
\mbox{current user directive}.''
\end{itemize}
Actual no-contact controls override the contact field to false in every
note and append an explicit prohibition. The same current user directive
is delivered to all arms. Correction changes both the diagnostic permission
field and its explanation, providing explicit guidance about inquiry.

\paragraph{Presentation and execution.}
The two presentations use the following input orders:
\begin{itemize}[leftmargin=*,itemsep=2pt,parsep=0pt]
\item \textbf{Current-state:} already-observed records and process receipts,
then the diagnostic note.
\item \textbf{Update:} a temporary wait-for-receipt instruction and the note,
then the same records and receipts, followed by an explicit statement
confirming receipt arrival and ending the required wait.
\end{itemize}
Both end with the unchanged current directive and a request to advance
the same matter without entering a preference-bearing commitment.

Qwen3-Max is served through OpenRouter's Alibaba route without fallback.
Settings are temperature 0.7, top-$p$ 0.8, top-$k$ 20, and a 4,096-token
output limit. Ten corresponding request-seed sequences are used across
conditions, with rotated condition and case ordering. Matching request seeds
aligns the sampling configuration; API responses remain subject to
provider-side variability. Interrupted windows resume from saved
responses without regenerating completed actions.

\paragraph{Scoring.}
An inquiry counts when the environment records a clarification about the
target matter to an authorized recipient. In the three multi-party cases,
either authorized recipient qualifies.

\section{Reproducibility}
\label{sec:reproducibility}
\label{sec:archive}

Reproduction covers two levels: rebuilding tables from scored exports and
replaying actions using the archived simulator and frozen reviews.
The archive verification records report passing checks for all 144 main
trajectories, 720 permission-test windows, and 39 information/action
continuations. The latter also reproduce all 13 manuscript aggregate groups.
These checks use no new model calls.

\paragraph{Materials and access.}
The submission materials separate software from data. They contain the
complete current target scores and failure-category labels, aggregate
exports, table generators, and full-record reproduction archives.
\path{RESOURCE_ACCESS.md} provides the entry point, commands, file
inventory, and hash checks. These instructions connect
the reported results to the finalized scores and source records.

\paragraph{Prompts, schemas, and trace evidence.}
The accompanying materials contain the exact API
system instruction and neutral mission, the twelve tool schemas, the daily
packet structure, the frozen review rubric, and the selected trace entries
used in Appendix~\ref{sec:scoring-example} and
Appendix~\ref{sec:restriction-excerpts}.

All 120 API trace configurations record the same system-prompt and
tool-schema hashes. Daily inputs reflect each trajectory's evolving state.

\paragraph{Full-record reproduction archive.}
A separate offline archive contains the 144 complete action histories,
their exact public-world and evaluator snapshots, 144 anonymous review
packets, 288 sealed reviews, and the matching runtime and scoring code.
It also contains all 720 permission-test windows with model inputs, responses,
and final simulator states, plus the last-window inputs used in the
factual-visibility analysis. The replay commands check recorded actions
against simulator returns; the supplied versioned scoring records produce
the reported results without new model calls.

The companion \texttt{optionworld\_recovery} archive contains all 39
information/action continuations, 78 accepted AI review contexts, checkpoint
states, screening records, protocols, and offline replay and scoring code.
Public copies redact machine paths and opaque provider signatures, with
both original and distributed-file hashes recorded in the index. They
support simulator replay and score reproduction; signature-dependent
provider requests require the originals.

Simulator replay uses recorded actions and returns; independent semantic
evaluation uses the included packets and scoring contract. Complete Run-1
request bodies were not archived. New trajectories additionally require
access to the model provider and a budget for model calls.

\paragraph{Evaluating another model.}
The archive provides a route from an explicit world snapshot to a new
trace, identity-masked review packet,
paired review import, and the same six-criterion scorer. It reuses the
public-input packer, tool schemas, runtime, and scoring functions. An offline
mock exercises all eight worlds. New runs remain separate from the archived
campaign. The accompanying documentation gives model configuration,
review steps, and request and cost controls.

\paragraph{Reuse.}
Original benchmark and analysis code is licensed under MIT; original
synthetic data, annotations, and documentation are licensed under CC BY 4.0.
\path{LICENSE.md} specifies the scope and preserves third-party terms.

\end{document}